\documentclass[a4paper,12pt]{elsarticle}

\usepackage{lineno,hyperref}
\modulolinenumbers[1]
\usepackage[T1]{fontenc}
\usepackage{graphicx}
\usepackage{amsmath,amssymb,amsfonts}
\usepackage{float, array, multirow, color, multirow, booktabs}
\usepackage{makecell}
\usepackage{hyperref}
\usepackage{setspace}
\usepackage{subfigure}

\makeatletter

\newcommand{\Rmnum}[1]{\expandafter\@slowromancap\romannumeral #1@}

\makeatother

\DeclareMathOperator*{\argmin}{argmin}

\begin{document}

\begin{frontmatter}



\title{PCNet: A Structure Similarity Enhancement Method for Multispectral and Multimodal Image Registration}


\author[label1,label2]{Si-Yuan~Cao}
\ead{karlcao@hotmail.com}

\author[label2]{Beinan Yu}
\ead{mr_vernon@hotmail.com}

\author[label2]{Lun Luo}
\ead{luolun@zju.edu.cn}

\author[label3]{Shu-Jie~Chen}
\ead{chenshujie@zjgsu.edu.cn}

\author[label2]{Chunguang~Li}
\ead{cgli@zju.edu.cn}

\author[label1,label2]{Hui-Liang~Shen\corref{correspondingauthor}}
\cortext[correspondingauthor]{Corresponding author}
\ead{shenhl@zju.edu.cn}

\address[label1]{Ningbo Innovation Center, Zhejiang University, Ningbo, 315100, Zhejiang, China}
\address[label2]{College of Information Science and Electronic Engineering, Zhejiang University, Hangzhou, 310027, Zhejiang, China}
\address[label3]{School of Computer and Information Engineering, Zhejiang Gongshang University, Hangzhou, 310018, Zhejiang, China}

\begin{abstract}
Multispectral and multimodal images are of important usage in the field of multi-source visual information fusion. Due to the alternation or movement of image devices, the acquired multispectral and multimodal images are usually misaligned, and hence image registration is pre-requisite. Different from the registration of common images, the registration of multispectral or multimodal images is a challenging problem due to the nonlinear variation of intensity and gradient. To cope with this challenge, we propose the phase congruency network (PCNet) to enhance the structure similarity of multispectral or multimodal images. The images can then be aligned using the similarity-enhanced feature maps produced by the network. PCNet is constructed under the inspiration of the well-known phase congruency. The network embeds the phase congruency prior into two simple trainable layers and series of modified learnable Gabor kernels. Thanks to the prior knowledge, once trained, PCNet is applicable on a variety of multispectral and multimodal data such as flash/no-flash and RGB/NIR images without additional further tuning. The prior also makes the network lightweight. The trainable parameters of PCNet is 2400$\times$ less than the deep-learning registration method DHN, while its registration performance surpasses DHN. Experimental results validate that PCNet outperforms current state-of-the-art conventional multimodal registration algorithms. Besides, PCNet can act as a complementary part of the deep-learning registration methods, which significantly boosts their registration accuracy. The percentage of the number of images under 1 pixel average corner error (ACE) of UDHN is raised from 0.2\% to 89.9\% after the processing of PCNet.

\end{abstract}




\end{frontmatter}


\section{Introduction}\label{sec:intro}
Multispectral and multimodal images, such as RGB and near-infrared (NIR) images \cite{brown2011multi}, flash/no-flash images \cite{he2014saliency}, and multispectral band images \cite{yasuma2010generalized,chakrabarti2011statistics}, usually contain much richer information compared to conventional RGB images. They are important data for multi-source visual information fusion applications including pedestrian detection and re-identification\cite{guan2019fusion, cao2022locality,an2022pedestrian}, image fusion \cite{muller2007robust,liu2018deep,dimitri2022multimodal}, and image denoising/dehazing \cite{xu2022model,guo2022image}. Pixel-level alignment of multispectral/multimodal images is a fundamental requirement for these tasks. Nevertheless, multispectral/multimodal images are prone to be misaligned due to the alternation or movement of image devices \cite{shen2014multi,yasuma2010generalized}. Therefore, registering multispectral/multimodal images is the primary problem for further computer vision and computational photography tasks. 

The most intractable aspect of multispectral/multimodal image registration is the ubiquitous variation of intensity and gradient among data from different sources \cite{chen2017normalized,shen2014multi}. To cope with the challenge, various image registration methods have been proposed. The registration methods can be categorized into feature-based ones and intensity-based ones \cite{zitova2003image}. Feature-based methods align images through feature detection, feature description, and transform estimation \cite{jiang2021review}. Intensity-based methods align images by finding the best correspondence maximizing (or minimizing) specific similarity measures between the two input images\cite{zimmer2019multimodal}. Various similarity measures have been specially developed for multispectral/multimodal images, however, they are usually of complicated forms, making their optimization problematic and computationally expensive \cite{maes1997multimodality}. Another kind of intensity-based method aims to enhance the structural similarity between images by some transforms in advance and then register the transformed images using common measures such as sum of squared differences (SSD) \cite{wachinger2010structural, cao2020boosting}. These common measures can be solved efficiently and effectively using concise optimizers such as classical gradient descent.

In this work, we propose the phase congruency network (PCNet) as a similarity enhancement method for multispectral/multimodal image registration. PCNet is a trainable network constructed under the guidance of phase congruency theory. Phase congruency theory \cite{morrone1987feature,kovesi2000phase} is a method of feature perception in the images which is invariant under illumination and contrast change. It has been employed in several tasks including similarity assessment \cite{liu2007phase,zhang2011fsim}, feature detection/extraction \cite{wang2011image,yuan2005iris}, image fusion \cite{bhatnagar2013directive, li2016infrared}, and image registration \cite{wong2007arrsi, ye2017robust,li2019rift}.

PCNet detects edge structures by estimating the phase congruency of different scales of frequency components. The magnitude of the output feature maps of PCNet merely depends on the congruency of phase among different frequency scales. Therefore, the similarity of the output features will be significantly enhanced in spite of the nonlinear variation of intensity and gradient. According to our experiments, PCNet outperforms state-of-the-art similarity enhancement methods and feature-based methods. As a trainable architecture, PCNet is regularized by our proposed modified learnable Gabor filters and the 2 trainable layers based on phase congruency theory, therefore, PCNet has relatively strong generalization ability. PCNet is only trained once on a subset of the CAVE multispectral data but performs well on several multispectral and multimodal datasets. The priors that regularize PCNet also make the network compact. The trainable parameter of PCNet is 2400$\times$ less than the deep-learning registration method DHN \cite{detone2016deep}, while the registration performance of PCNet surpasses DHN. What's more, PCNet can act as a complementary part of the deep-learning registration methods such as DHN, MHN \cite{le2020deep} and UDHN \cite{zhang2020content}, which can significantly boost their registration accuracy. For example, the percentage of the number of images under 1 pixel average corner error (ACE) of UDHN is raised from 0.2\% to 89.9\% after the processing of PCNet. It is worth noting that as is explored in \cite{le2020deep}, a more combination of the CNN architectures can't guarantee such accuracy.

The multispectral/multimodal image registration procedure using PCNet is performed as follows. First, we build the phase congruency architecture under the guidance of phase congruency theory, which contains 2 trainable layers. Second, we employ the modified learnable Gabor kernels for multi-scale frequency component extraction, which significantly reduce the difficulty of network training and improve the generalization ability at the same time. Third, we introduce a normalized structural similarity loss for the unsupervised training of PCNet. Finally, we present the image registration framework using the image pyramid and gradient descent algorithm. 

To summarize, the main contributions of this work are as follows: 
\begin{itemize}
\item Based on phase congruency, we propose the phase congruency network (PCNet) that can significantly improve the structural similarity between images having nonlinear variation of intensity and gradient. The network contains two parts, i.e., the phase congruency architecture and modified learnable Gabor kernels.
\item To cope with the problem that no ground truth exists for phase congruency features, we design a normalized structural similarity loss depicting the similarity between the output feature maps of PCNet. The network can then be trained in an unsupervised manner.
\item We show that thanks to the prior knowledge of phase congruency, PCNet can be trained once on multispectral data but works effectively on a wide range of multispectral/multimodal data. PCNet can not only outperform state-of-the-art registration methods but also works as a complementary part of the deep-learning methods which significantly boost their registration accuracy.
\end{itemize}

\section{Related Work}\label{sec:relatedwork}
The registration techniques for multispectral/multimodal images can be coarsely categorized into intensity-based methods and feature-based methods. In the following, we will present a brief review of the similarity enhancement intensity-based methods. For more detailed surveys on image registration, please refer to \cite{jiang2021review, ma2020image}.

For intensity-based methods, we focus on the similarity enhancement ones as they are generally more efficient and easy to optimize compared to the conventional complicated multispectral/multimodal measures such as mutual information (MI) \cite{maes1997multimodality}, robust selective normalized cross correlation (RSNCC) \cite{shen2014multi}, and residual correlation ratio (CR) \cite{hel2013matching}.

Entropy image (EI) \cite{wachinger2010structural} obtains the consistent structure using the Shannon entropy. EI adopts local histogram to compute the probability of different intensity levels and then computes the entropy as the enhanced structure.

WLD \cite{chen2009wld} detects local texture information based on Weber's law \cite{fechner1966elements}. The law states that for the human visual system, the perceived change in stimuli is proportional to the initial stimuli. In \cite{yang2013two}, the differential excitation component of WLD is employed for multimodal image registration.

Structure consistency boosting (SCB) transform \cite{cao2020boosting} adopts the statistical prior from natural images. The consistent inherent edge structures are enhanced using a transform depicted by generalized Gaussian distribution (GGD) with 3 trainable parameters.

Dense adaptive self-correlation (DASC) \cite{kim2016dasc} descriptor is also a similarity enhancement algorithm leveraging machine learning. The algorithm is established based on the local self-similarity prior. The similarity between patches in a local support window is computed and encoded into a descriptor. The optimal sampling pattern for patch similarity computation is learned using the support vector machine (SVM).

Several feature-based methods have been proposed for registering multispectral/multimodal images. Log-Gabor histogram descriptor (LGHD) \cite{aguilera2015lghd} and radiation-variation insensitive feature transform (RIFT) \cite{li2019rift} employ the log-Gabor filter to alleviate the nonlinear variation of intensity and gradient. LGHD builds a histogram for the filter response of each scale of the log-Gabor filter. The combined histogram of different scales finally forms a descriptor. Different from LGHD, RIFT does the summation over each scale of the log-Gabor filter and then builds a maximum index map (MIM) by recording the channel number of the maximum value through different orientations of the log-Gabor filter responses. A histogram similar to SIFT is established based on the MIM and is used as the final descriptor. 

Recently, many deep-learning methods have been proposed to register unaligned images. Deep homography network (DHN) \cite{detone2016deep} adopts a VGG-style network to directly predict the global motion between the concatenated source and target images. MHN \cite{le2020deep} is then proposed to align image pairs by cascading 3 levels of VGG-style networks, which significantly improves the registration accuracy. Another network based on ResNet-34 \cite{he2016deep} is also proposed to align images with moving foregrounds, named UDHN \cite{zhang2020content}. 

\section{Phase Congruency Theory}

\begin{figure}[!tb]
\centering
\includegraphics[scale=0.7]{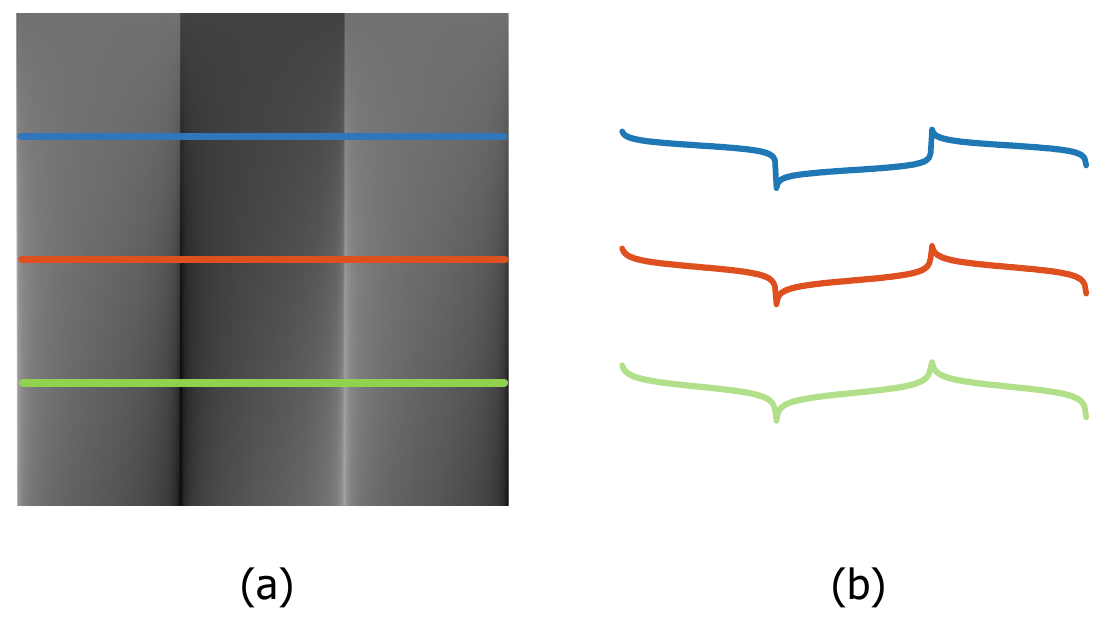}
\caption{The constructed grating and its corresponding profiles. (a) the grating constructed using (\ref{eq:wave_series}). (b) the corresponding profiles from (a).}
\label{fig:PC_plot}
\end{figure}

In an image, edge features are perceived where the different scales of wavelets meet maximum phase congruency. For better illustration, we construct a grating in Fig. \ref{fig:PC_plot} (a) using the series
\begin{equation}\label{eq:wave_series}
\sum_{s=0}^{\infty} \frac{1}{2s+1}\sin((2s+1)x + \phi),
\end{equation}
where $x$ varies along the horizontal direction, and $\phi$ the vertical, denoting the congruent phase shift. $s$ denotes the scale of the wavelet series. Profiles of different phase shifts are drawn in Fig. \ref{fig:PC_plot} (b). We observe that the congruency of phase at every phase shift produces a clearly perceived feature, namely the edge structure of an image.

Different practical ways for calculating phase congruency have been proposed in \cite{morrone1987feature,morrone1995adaptive,robbins19972d,kovesi2000phase}, and a robust approach was introduced in \cite{kovesi2000phase}. Concretely, the phase congruency feature is computed by first convolving the image with a quadrature pair of wavelet filters at each scale $s$. The wavelet filters contain the even-symmetric and odd-symmetric parts. The frequency responses of the wavelet filters at each scale are denoted by $e_s(x,y)$ for the even and $o_s(x,y)$ the odd. We can then compute the amplitude of response at each scale
\begin{equation}\label{eq:A_s}
A_{s} (x, y) = \sqrt {e_{s} (x, y)^{2} + o_{s} (x, y)^{2}},
\end{equation}
and the phase
\begin{equation}\label{eq:phi_s}
\phi _{s} (x, y) = \arctan(o_{s}(x, y)/e_{s} (x, y)).
\end{equation}
Then the local energy can be computed as 
\begin{equation}\label{eq:E_s}
E (x, y) = \sqrt {(\sum_s e_{s} (x, y))^2 
                     +(\sum_s o_{s} (x, y))^2},
\end{equation}
and its corresponding phase as 
\begin{equation}\label{eq:phibar_s}
\bar{\phi} (x, y) = \arctan(\sum_s o_{s} (x, y)/\sum_s e_{s} (x, y)).
\end{equation}
For clarity, we refer to the phase of multi-scale filter response as phase (denoted by $\phi_{s} (x, y)$), and the phase of local energy as mean phase (denoted by $\bar{\phi} (x, y)$) in the following.

The phase congruency at pixel position $(x,y)$ is measured by the ratio of the local energy and the accumulation over scales of the amplitude of response 
\begin{equation}\label{eq:PC_0}
\begin{aligned}
PC_0(x,y)& = \frac {{{{E (x, y) }}}}{\sum_{s} {{A_{s}(x,y)} +\xi }}\\
  &=\frac {\sum_{s} {{{{A_{s}(x,y)\cos(\phi_s(x,y)-\bar{\phi}(x,y))} }}}}
{\sum_{s} {{A_{s}(x,y)} +\xi}},
\end{aligned}
\end{equation}
where $\xi$ is a small value that avoids division by zero. It is claimed that the above phase deviation estimation calculation is prone to produce blurry phase congruency features \cite{kovesi2000phase}. To endow the phase deviation estimation with more ocular discriminability, a correction term is added in. The phase congruency is then formulated as
\begin{equation}\label{eq:PC_1}
\begin{aligned}
PC_1(x,y) = \frac{
\sum_{s} {{{{A_{s}(\cos(\phi_s-\bar{\phi})-|\sin(\phi_s-\bar{\phi})|)} }}}}
{\sum_{s} {{A_{s}} +\xi}},
\end{aligned}
\end{equation}
where the pixel position $(x,y)$ is omitted for notation simplification.  

The features extracted by the phase congruency procedure can weaken the inconsistency of the original image edge. As a similarity enhancement technique, phase congruency theory has been adopted in many multispectral/multimodal image registration methods, such as ARRSI \cite{wong2007arrsi}, HOPC \cite{ye2017robust}, and the above-mentioned RIFT \cite{li2019rift}. ARRSI detects feature points and performs normalized cross-correlation (NCC) \cite{yoo2009fast} matching on the maximum moment map of phase congruency. HOPC combines the orientation and the amplitude of phase congruency with the histograms of oriented gradients (HOG) descriptor to construct a similarity enhanced feature descriptor. RIFT employs the maximum moment map of phase congruency to conduct feature detection. Different from the above registration methods that directly use the product of phase congruency computation \cite{kovesi2000phase}, we take phase congruency as a powerful prior knowledge to design a trainable network that can enhance the similarity of multispectral/multimodal images.

\section{Phase Congruency Network for Structural Similarity Enhancement}

\begin{figure*}[!tb]
\centering
\includegraphics[scale=0.225]{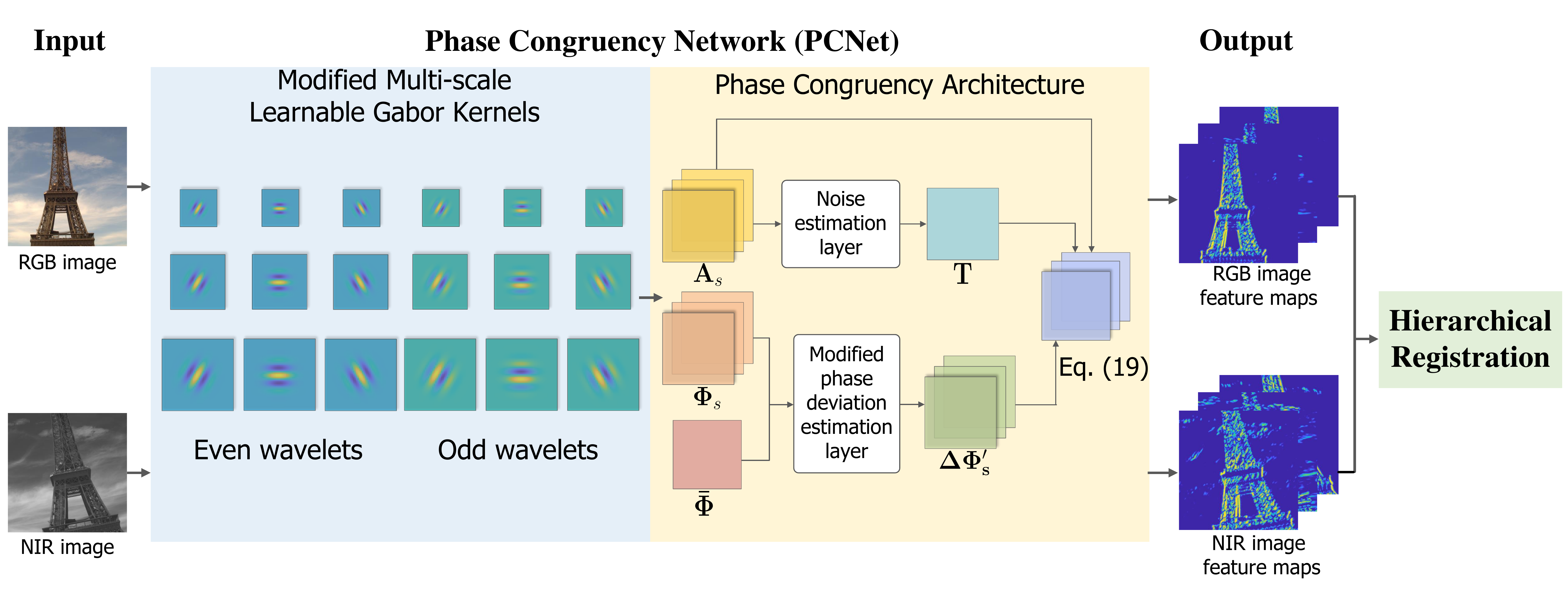}
\caption{The schematic diagram of phase congruency network (PCNet). The network is constructed based the phase congruency. Two trainable layers (noise estimation layer and modified phase deviation estimation layer) are learnable parts of the phase congruency architecture. The modified learnable Gabor kernels are employed as the quadrature wavelet bank satisfying phase congruency theory. The input RGB and NIR images are processed by PCNet and then transformed into similarity-enhanced feature maps. The feature maps are then fed into the hierarchical registration module. {Note that, for illustration, the diagram shows the phase congruency architecture within one orientation, and shows the similarity enhancement in the inference stage.}}
\label{fig:PCNet_scheme}
\end{figure*}

For multispectral and multimodal images, the phase congruency procedure can enhance structure consistency regardless of the nonlinear variation of image intensity and gradient. However, the present phase congruency procedure is oversensitive to image noise and fake edges. To produce satisfactory similarity enhancement results, we proposed the phase congruency network (PCNet).

Our PCNet is constructed based on phase congruency \cite{kovesi2000phase}. Several modification strategies are put forward for network construction. Fig. \ref{fig:PCNet_scheme} depicts the schematic diagram of PCNet. The input images are first convoluted by the modified learnable Gabor kernels and then fed into the phase congruency architecture. The similarity enhanced structure outputs are finally produced by PCNet. {Note that Fig. \ref{fig:PCNet_scheme} illustrates the inference stage of PCNet, in which the two input images are not aligned. For the training stage please kindly refer to Fig. \ref{fig:PCNet_train}.} The modified learnable Gabor kernels consist of the learnable convolutional kernels and the fixed Gabor wavelets. The Gabor wavelets contain quadrature pairs of wavelets, and are steerable and scalable, satisfying the phase congruency theory. The phase congruency architecture is constructed based on the aforementioned theory with two trainable layers, namely noise estimation layer and modified phase deviation estimation layer. We will describe the details of PCNet below.

We first reformulate (\ref{eq:PC_1}) into a more compact form for tensor manipulation,

\begin{equation}\label{PCNet0}
\begin{aligned}
\mathbf{P} = (\sum\limits_{s} \mathbf{A}_s \circ \Delta \mathbf{\Phi}_s) \oslash (\sum\limits_{s} \mathbf{A}_s + \xi \cdot \mathbf{1}),
\end{aligned}
\end{equation}
where $\oslash$ denotes pointwise division or Hadamard division\cite{cyganek2013object}. $\mathbf{1} \in \mathbb{R}^{H\times W}$ is an all-ones matrix. $\xi \cdot \mathbf{1}$ prevents the numerator from being divided by zero and will be omitted in the following. $\mathbf{P} \in \mathbb{R}^{H\times W}$, $\mathbf{A}_s \in \mathbb{R}^{H\times W \times S}$, and $\Delta \mathbf{\Phi}_s \in \mathbb{R}^{H\times W \times S}$, with $S$ being the amount of filter scales. $\Delta \mathbf{\Phi}_s$ indicates the phase deviation estimation layer, defined as
\begin{equation}\label{eq:phibarM_s_Kovesi}
\Delta \mathbf{\Phi}_s = \cos(\mathbf{\Phi}_s-\bar{\mathbf{\Phi}}) - |\sin(\mathbf{\Phi}_s-\bar{\mathbf{\Phi}})|,
\end{equation}
where $\mathbf{\Phi}_s$ denotes the phase map and $\bar{\mathbf{\Phi}}$ the mean phase map.

To obtain satisfactory similarity enhancement results for image registration, two trainable layers are proposed and then combined into PCNet.

\subsection{Noise Estimation Layer}

The computation of phase congruency leveraging (\ref{PCNet0}) is sensitive to noise because in the natural image noise forms small edges. Hence, the noise should be estimated and eliminated before computing the phase congruency. As illustrated previously, the local energy is the square root of two independent random variables, each following a standard normal distribution. Thus the noise of the local energy will have a Rayleigh distribution. We denote $\mathbf{M}_{\mathrm{R}}$ as the mean of the Rayleigh distribution, which is computed as
\begin{equation}\label{eq:M_R}
\mathbf{M}_{\mathrm{R}} = \sqrt{\frac{\pi}{2}} \cdot 
\mathbf{V}_{\mathrm{G}},
\end{equation}
where $\mathbf{V}_{\mathrm{R}}$ denotes the variance of the Rayleigh distribution computed as
\begin{equation}\label{eq:V_R}
\mathbf{V}_{\mathrm{R}} = \sqrt{\frac{4 - \pi}{2}} \cdot 
\mathbf{V}_{\mathrm{G}}.
\end{equation}
The Rayleigh noise map can then be estimated as
\begin{equation}\label{eq:T}
\mathbf{{T}} = \mathbf{M}_{\mathrm{R}} +  \mathbf{V}_{\mathrm{R}}.
\end{equation}
$\mathbf{V}_{\mathrm{G}}$ in (\ref{eq:M_R}) and (\ref{eq:V_R}) denotes the variance of the end position of the local energy vector, which is estimated as
\begin{equation}\label{eq:V_G}
\mathbf{V}_{\mathrm{G}} = \tau \cdot (\mathbf{1} - (\mathbf{1}/{\alpha})^{\circ N_s})
\oslash(\mathbf{1}-{\mathbf{1}}/{\alpha} + \xi \cdot \mathbf{1}),
\end{equation}
where $N_s$ denotes the number of the frequency scales and $^{\circ}$ denotes the pointwise power or Hadamard root \cite{reams1999hadamard}. $\tau$ can be directly estimated from the local energy. Note that the original $\alpha$ here is defined as the scaling factor between successive filters, but in this work, we make it a trainable unit as it directly controls the overall noise threshold. Unit $\alpha$ will be updated by the gradient flow in the training stage, making the Rayleigh noise map into the noise estimation layer of our PCNet.

Once having the threshold produced by the noise estimation layer, the noise can be removed by soft thresholding
\begin{equation}\label{eq:soft_thre}
f(m) = \left\{ \begin{aligned}
&m - t &\mathrm{if} &\ m>t\\
&0 &\mathrm{if} &\ m\leq t\\  
 \end{aligned}
\right.,
\end{equation}
where $t$ denotes the estimated threshold. For our PCNet, the soft thresholding operation can be perfectly modified into the rectified linear units \cite{nair2010rectified}, namely the $\mathrm{ReLU}$ activation function, which can then be formulated as 

\begin{equation}\label{eq:PCNet1}
\mathbf{P} = \mathrm{ReLU}( \sum\limits_{s} \mathbf{A}_s \circ \Delta \mathbf{\Phi}_s - \mathbf{T}) \oslash \sum\limits_{s} \mathbf{A}_s.
\end{equation}

\subsection{Modified Phase Deviation Estimation Layer}

The original phase deviation estimation layer according to \cite{kovesi2000phase} is formed by (\ref{eq:phibarM_s_Kovesi}). The added correction term can make the output phase congruency features visually thinner. However, it is not clear that ocular discriminability is beneficial to structure similarity enhancement. Hence, we employ a trainable term $\beta$ in the phase deviation estimation layer, yielding the modified phase deviation estimation layer
\begin{equation}\label{eq:phibarM'_s}
\Delta \mathbf{\Phi}'_s = \cos(\mathbf{\Phi}_s-\bar{\mathbf{\Phi}}) - \beta\cdot|\sin(\mathbf{\Phi}_s-\bar{\mathbf{\Phi}})|.
\end{equation}

By adopting the above trainable layers, our PCNet has the mathematical form
\begin{equation}\label{eq:PCNet}
\begin{aligned}
\mathbf{P}(\alpha,\beta) = \mathrm{ReLU}(\sum\limits_{s} \mathbf{A}_s \circ \Delta \mathbf{\Phi}'_s - \mathbf{T}) 
\oslash \sum\limits_{s} \mathbf{A}_s.
\end{aligned}
\end{equation}

With the assistance of phase congruency, we have constructed a network architecture with trainable units. However, we still need a proper set of wavelet filters for multi-scale frequency component extraction. 

\subsection{Modified Multi-scale Learnable Gabor Kernels}

It is a straightforward idea to construct a series of convolutional kernels for multi-scale frequency component extraction. However, there are no ground truth phase congruency feature maps for network training, and hence it is difficult to train the convolutional kernels without regularization. Furthermore, the large-scale frequency components of phase congruency generally require relatively huge convolutional kernel sizes (e.g. a kernel size of $25 \times 25$), which also increases the difficulty of kernel training.

\begin{figure}[!tb]
\centering
\includegraphics[scale=0.3]{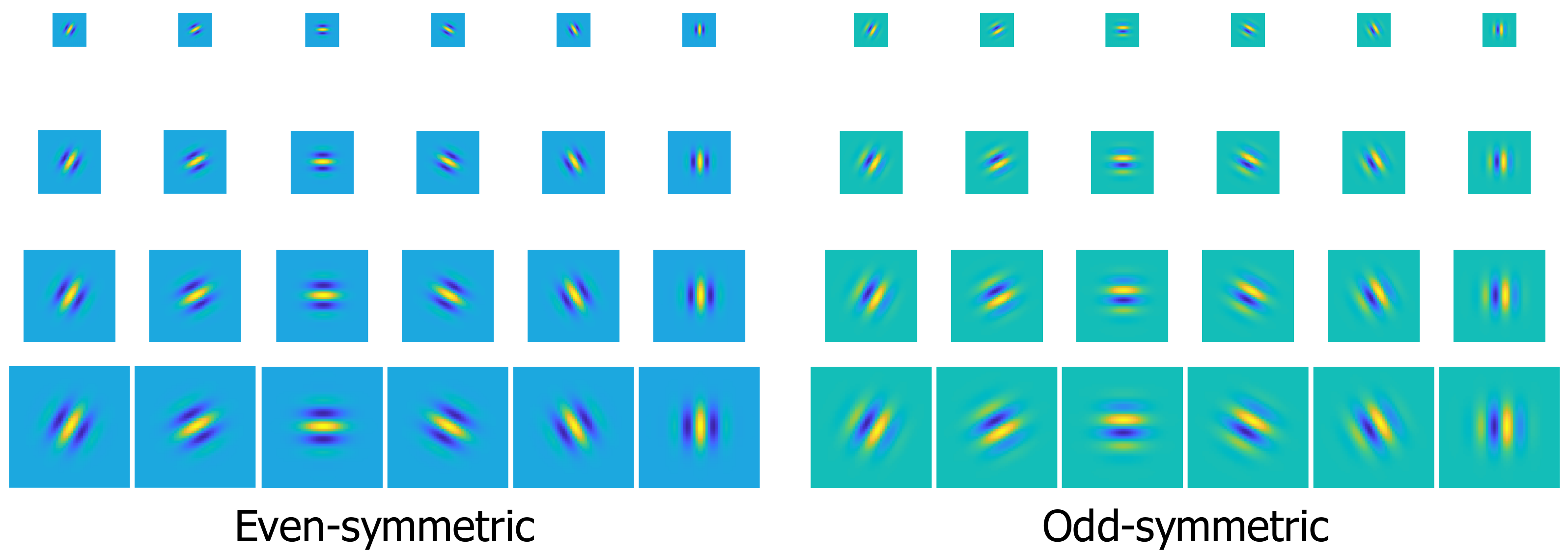}
\caption{Examples of the steerable and scalable Gabor wavelets.}
\label{fig:gabor_wavelets}
\end{figure}

Considering the above issues, we adopt modified multi-scale learnable Gabor kernels as part of the network architecture for multi-scale frequency component extraction. As illustrated in Fig. \ref{fig:gabor_wavelets}, the Gabor wavelets are steerable and scalable filters, which is created by Dennis Gabor \cite{gabor1946theory}. It is claimed that simple cells in the visual cortex of mammalian brains can be modeled by the Gabor functions \cite{marcelja1980mathematical}, thus the Gabor wavelets are thought to be similar to the perception of the human visual system. What's more, it has been shown that the shallow layers of image-trained CNNs tend to learn filters resembling Gabor filters \cite{luan2018gabor}. Gabor filters are composed of a pairwise bank of multi-scale quadrature wavelets, which perfectly satisfies the requirement of the phase congruency theory. The Gabor wavelets are defined as 

\begin{equation}\label{eq:Gabor_wavelets}
G_{u,v}(\mathbf{z})
=\frac{{\|{\mathbf{k}}_{u,v}\|^2}}{\sigma^2}
\textrm{e}^{\left(-\|{\mathbf{k}}_{u,v}\|^{2}\|{\mathbf{z}}\|^{2}/2\sigma^{2}\right)}
\bigl(\textrm{e}^{i{\mathbf{k}}_{u,v}^\mathsf{T}{\mathbf{z}}}-\textrm{e}^{-\sigma^{2}/2}\bigr).
\end{equation}

\begin{figure}[!tb]
\centering
\includegraphics[scale=0.35]{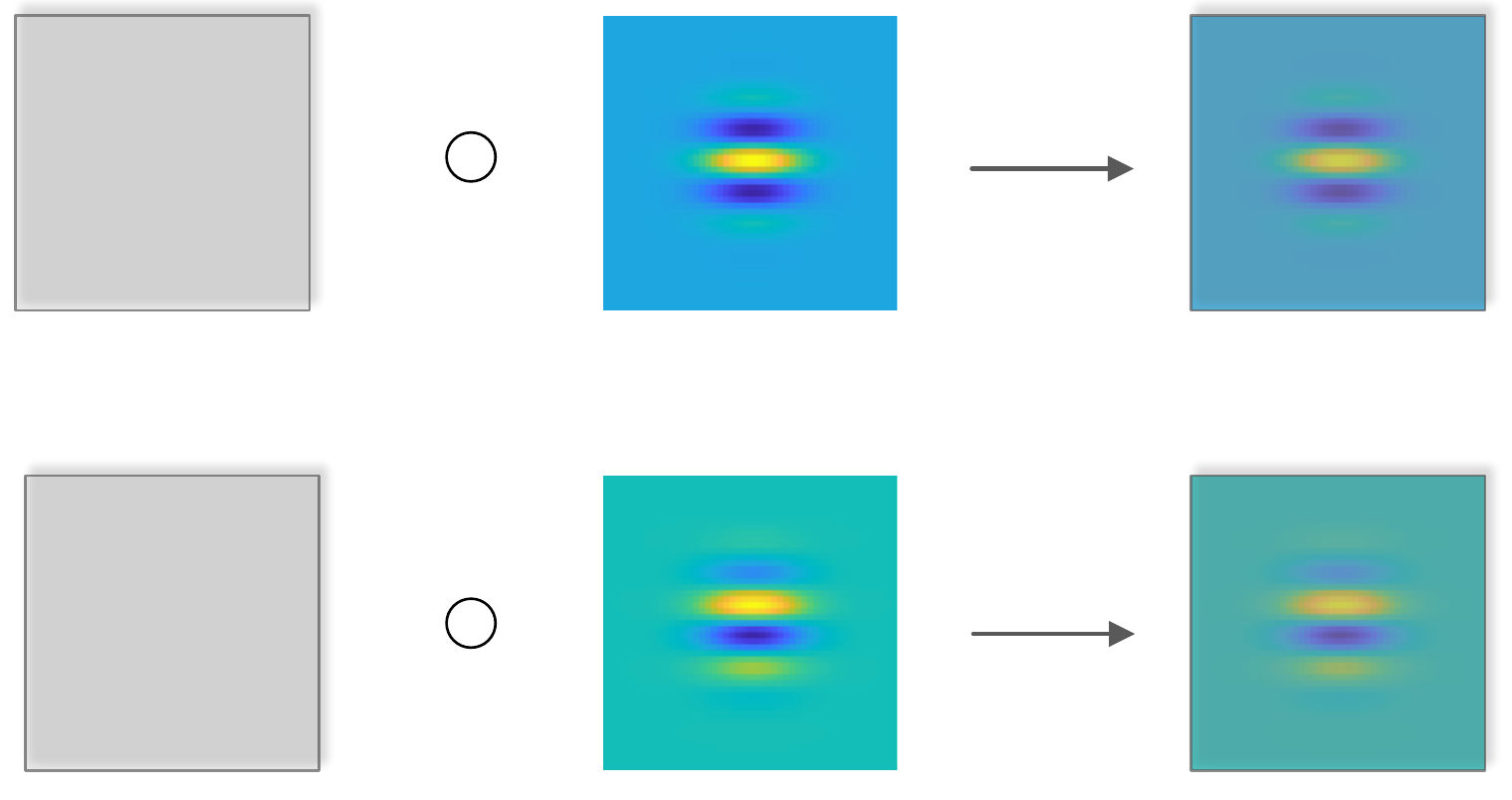}
\caption{Illustration of the learnable Gabor filters.}
\label{fig:learnable_gabor_filter}
\end{figure}

The Gabor filters are made learnable by making pointwise production with the learnable convolutional kernels having the same size \cite{luan2018gabor} as illustrated in Fig. \ref{fig:learnable_gabor_filter}. The original purpose of introducing Gabor wavelets into CNN architecture in \cite{luan2018gabor} is to guide the learnable convolutional kernels with directional information, which lightens the image classification deep networks and improves classification performance. On the contrary, in our work, the modulation provides the CNN layer with more constraints, which significantly reduces the difficulty of network training. It is worth noting that compared to \cite{luan2018gabor}, our modified learnable Gabor kernels are markedly different in 3 aspects:
\begin{itemize}
\item We employ the full parts (even-symmetric and odd-symmetric) of Gabor wavelets in accordance with the phase congruency theory, while \cite{luan2018gabor} only uses the even-symmetric part to guide the direction of convolutional kernels for the classification purpose.
\item We achieve the multi-scale convolution by directly enlarging the size of kernels at different scales, which perfectly matches the requirement of phase congruency theory, whereas \cite{luan2018gabor} resizes the input features using the max-pooling operation.
\item Considering that the Gabor filters have the drawback of being over-sensitive to the DC component of the signal, we modify the Gabor wavelets by subtracting their corresponding averages. 

\end{itemize}

\begin{figure}[!tb]
\centering
\includegraphics[scale=0.4]{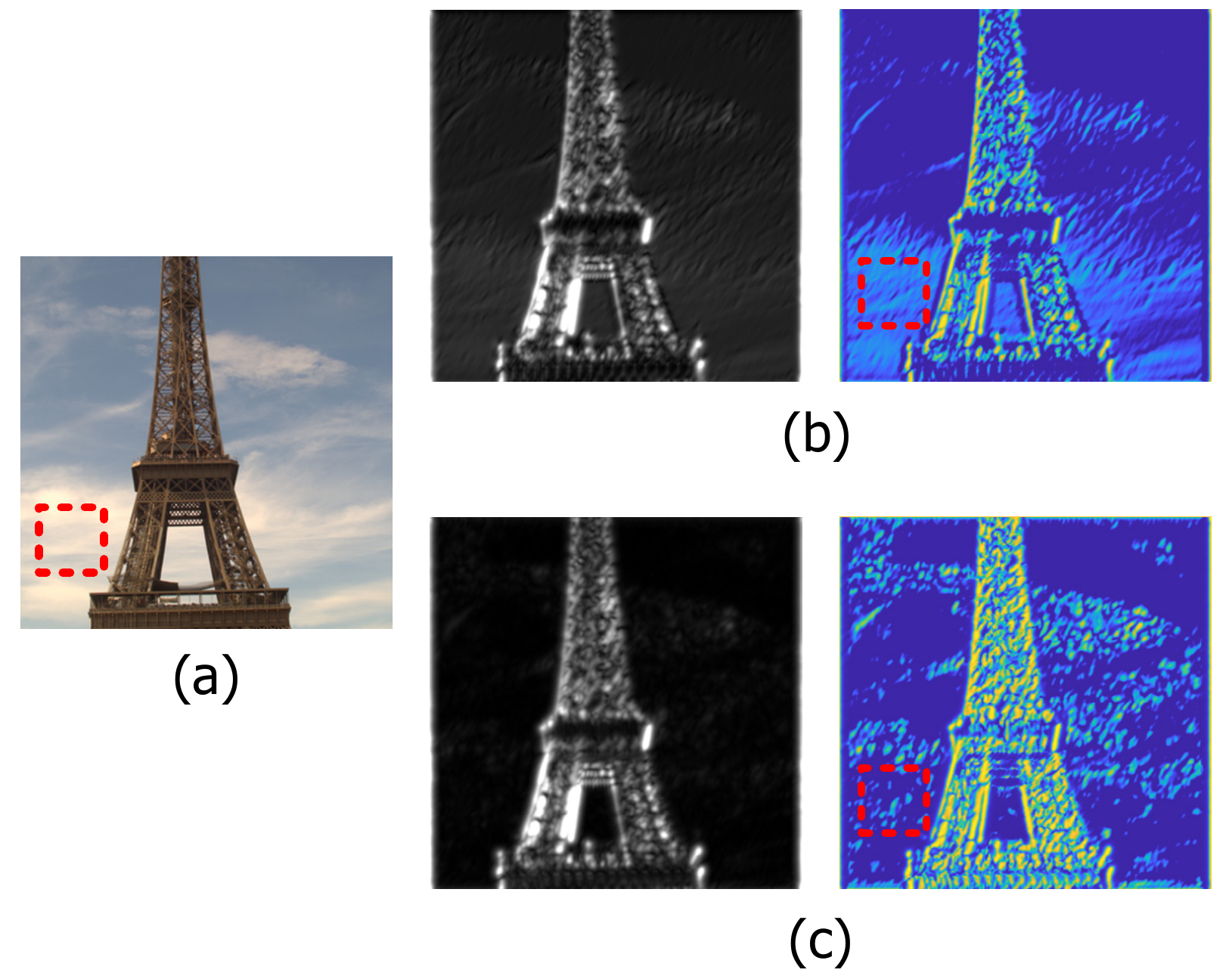}
\caption{The filter responses together with the phase congruency outputs for (b) the original Gabor filters and (c) the modified Gabor filters. The red boxes highlight the areas for detailed comparison.}
\label{fig:gabor_filter_mean}
\end{figure}

To illustrate the effectiveness of the modified Gabor filters, we draw the filter responses together with the phase congruency outputs for the original Gabor filters and the modified ones in Fig. \ref{fig:gabor_filter_mean}. It is observed that without the modification, the filter has an obvious response for the DC components of the image. As a result, the corresponding phase congruency output is also sensitive to the DC component. On the contrary, our modified Gabor filters significantly alleviate this problem.

For each orientation of the modified learnable Gabor kernels, the phase congruency architecture adopts the corresponding filter outputs and produces a channel of the phase congruency map. At last, channels of map form multi-channel phase congruency feature maps. The dimension of the features is determined by the orientation number of the modified learnable Gabor kernels. By denoting $o$ as the orientation, the calculation of our PCNet is finally formulated as
\begin{equation}\label{eq:PCNet_multiori}
\begin{aligned}
\mathbf{P}_o(\alpha,\beta, W) = \ & \mathrm{ReLU}( \sum\limits_{s} \mathbf{A}_{s,o} \circ \Delta \mathbf{\Phi}'_{s,o} - \mathbf{T}_o) \oslash \sum\limits_{s} \mathbf{A}_{s,o},
\end{aligned}
\end{equation}
where $W$ denotes the trainable parameters in the modified learnable Gabor kernels. 

\subsection{Normalized Structural Similarity Loss for Unsupervised Network Training}

Our PCNet can be trained by the stochastic gradient descent (SGD) algorithm using a proper loss function. However, as previously mentioned, the ground truth phase congruency feature maps do not exist for network training. In this work, we employ a loss function that directly compares the pairwise similarity of the output phase congruency feature maps of two input multispectral or multimodal images. In this way, an unsupervised learning framework is established, which perfectly matches our requirement of enhancing the structural similarity of the input images. 

The network is trained using the siamese strategy. Let $I_1$ and $I_2$ be two original band images, $\mathbf{P}_1$ and $\mathbf{P}_2$ be the outputs after applying our PCNet. Considering that the phase congruency features produced by PCNet are similar to edge structures, we employ SSIM to measure the similarity of $\mathbf{P}_1$ and $\mathbf{P}_2$,
\begin{equation}\label{eq:SSIM_loss}
L = 1 - \sum_o \mathrm{SSIM}(\mathbf{P}_{1,o},\mathbf{P}_{2,o})/N_o.
\end{equation}
Here comes a problem that the above loss function tends to produce an ambiguous guide for the network training. For example, a pair of outputs with all $0$ intensity will produce a minimized loss function with $0$ value. Consequently, the structural content of the image will be exterminated and the registration procedure will fail. To cope with this problem, we adopt the gradient of an image to protect its structural information, yielding the modified loss function
\begin{equation}\label{eq:norm_SSIM_loss}
L = \frac{1 - \sum \limits_o \mathrm{SSIM}(\mathbf{P}_{1,o},\mathbf{P}_{2,o})/N_o}{|\sum \limits_l \sum \limits_o (\|\nabla_l\mathbf{P}_{1,o}\|_1 + \|\nabla_l \mathbf{P}_{2,o}\|_1)/N_o|^{c}},
\end{equation}
where the operator $\nabla_l$, $l\in\{x,y\}$, represents the gradient computation along the horizontal and vertical directions. $c$ is the structure protection parameter that balances the degree of similarity enhancement and structure protection. The networks can be trained using the SGD algorithm.

\subsection{Hierarchical Motion Estimation Based Image Registration}

In this work, we adopt the intensity-based algorithm for image registration. The intensity-based method registers the misaligned images by minimizing (or maximizing) similarity measures with reference to the parametric or non-parametric transforms. We employ the sum of squared differences (SSD) as the registration measure. It has been widely used for a variety of image registration tasks\cite{szeliski2007image}. The registration procedure is formulated as
\begin{equation}\label{eq:ssd_reg}
\mathbf{\hat{a}} = \argmin_\mathbf{a} \sum_{p\in\Omega(\mathbf{a}) } (I_F(p,\mathbf{a})-I_R(p))^2,
\end{equation}
where $\mathbf{a}$ denotes the registration parameter for correcting the inconsistency of the coordinate between the reference image $I_R$ and floating image $I_F$, $\Omega(\mathbf{a})$ denotes the meaningful overlapping area of the warped $I_F$. We focus on the affine transform for modeling the registration parameters in this work. The affine transform can handle image deformation such as rotation, scaling, translation, shearing, and any combinations of them \cite{abbasi1999shape}. As a parametric transform, it has been widely adopted for multispectral and multimodal image registration \cite{ klein2012multispectral}. 

We employ the Gaussian pyramid for hierarchical motion parameter estimation \cite{szeliski2007image}. The registration parameters in each layer are updated using gradient descent optimization
\begin{equation}\label{eq:img_pyramid_updating}
\mathbf{a}^{t+1} = \mathbf{a}^t - \eta\nabla_\mathbf{a}J(I_R, I_F, \mathbf{a}^t),
\end{equation}
where $\eta$ denotes the step size, $t$ the iteration number during the optimization, and $\nabla_\mathbf{a}J(I_R, I_F, \mathbf{a}^t)$ the gradient of (\ref{eq:ssd_reg}) with respect to $\mathbf{a}^t$.


\section{Experiments}\label{sec:experiments}
\begin{figure}[!tb]
\centering
\includegraphics[scale=0.7]{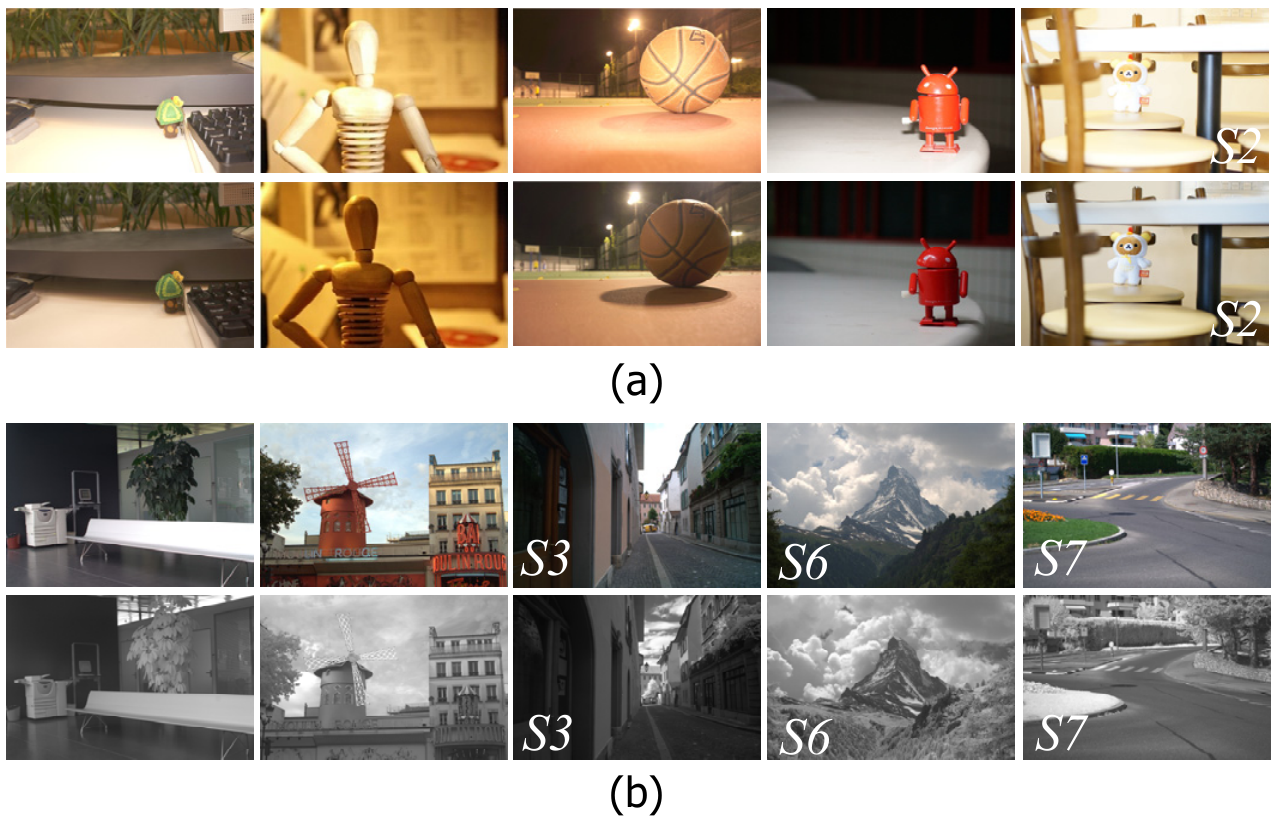}
\caption{Example images for RGB/NIR and flash/no-flash datasets. (a) RGB/NIR dataset (b) Flash/no-flash dataset. The scenes to be used for illustration are marked with \textit{S2}, \textit{S3}, \textit{S6}, and \textit{S7}.}
\label{fig:rgbnir_flf_exp}
\end{figure}

\begin{figure*}
    \centering
    \subfigure[Evaluation on CAVE dataset.]
    {
        \centering
        \includegraphics[scale=0.34]{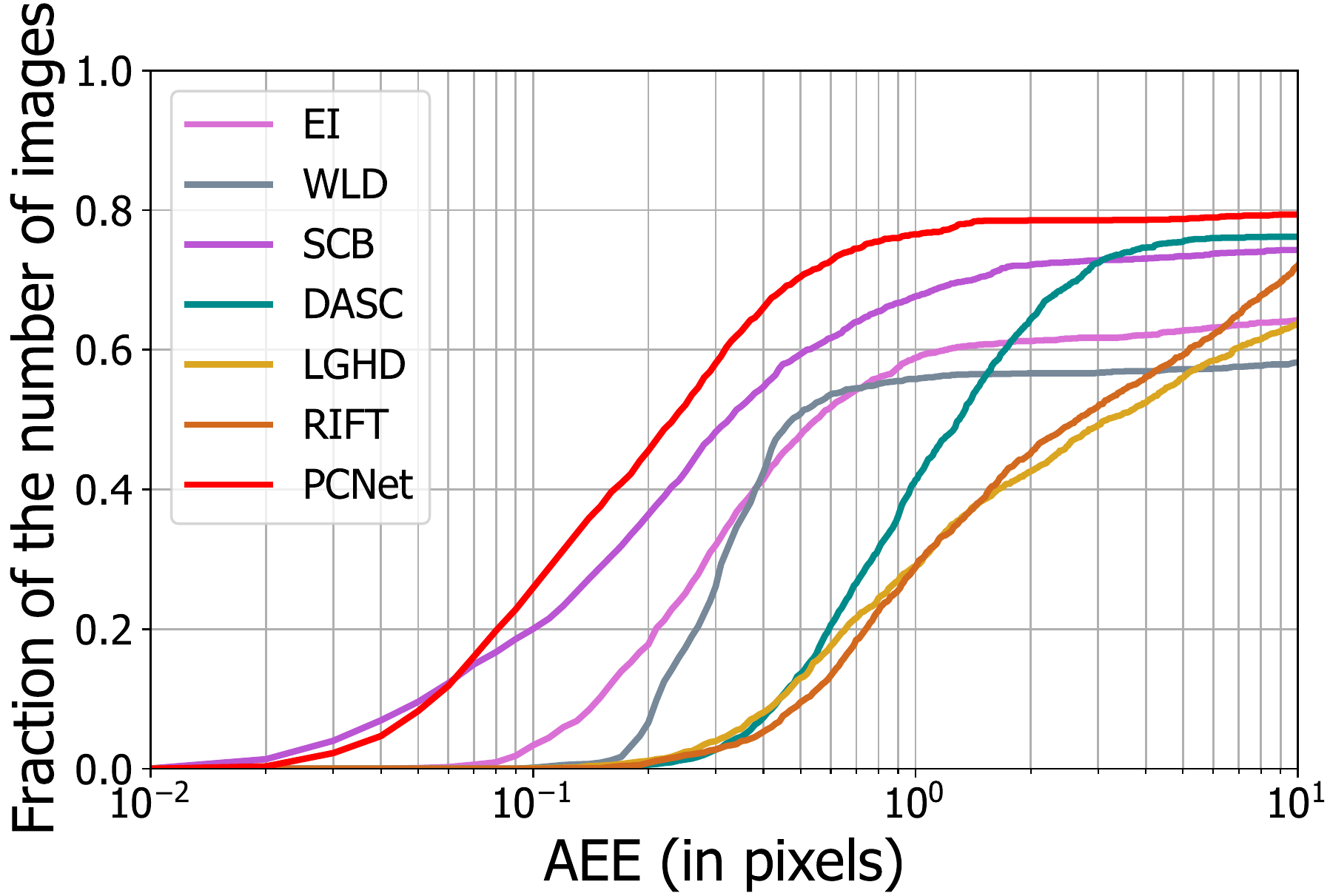}
        \label{fig:cave_conventinonal}
    }
    \subfigure[Evaluation on Harvard dataset.]
    {
        \centering
        \includegraphics[scale=0.34]{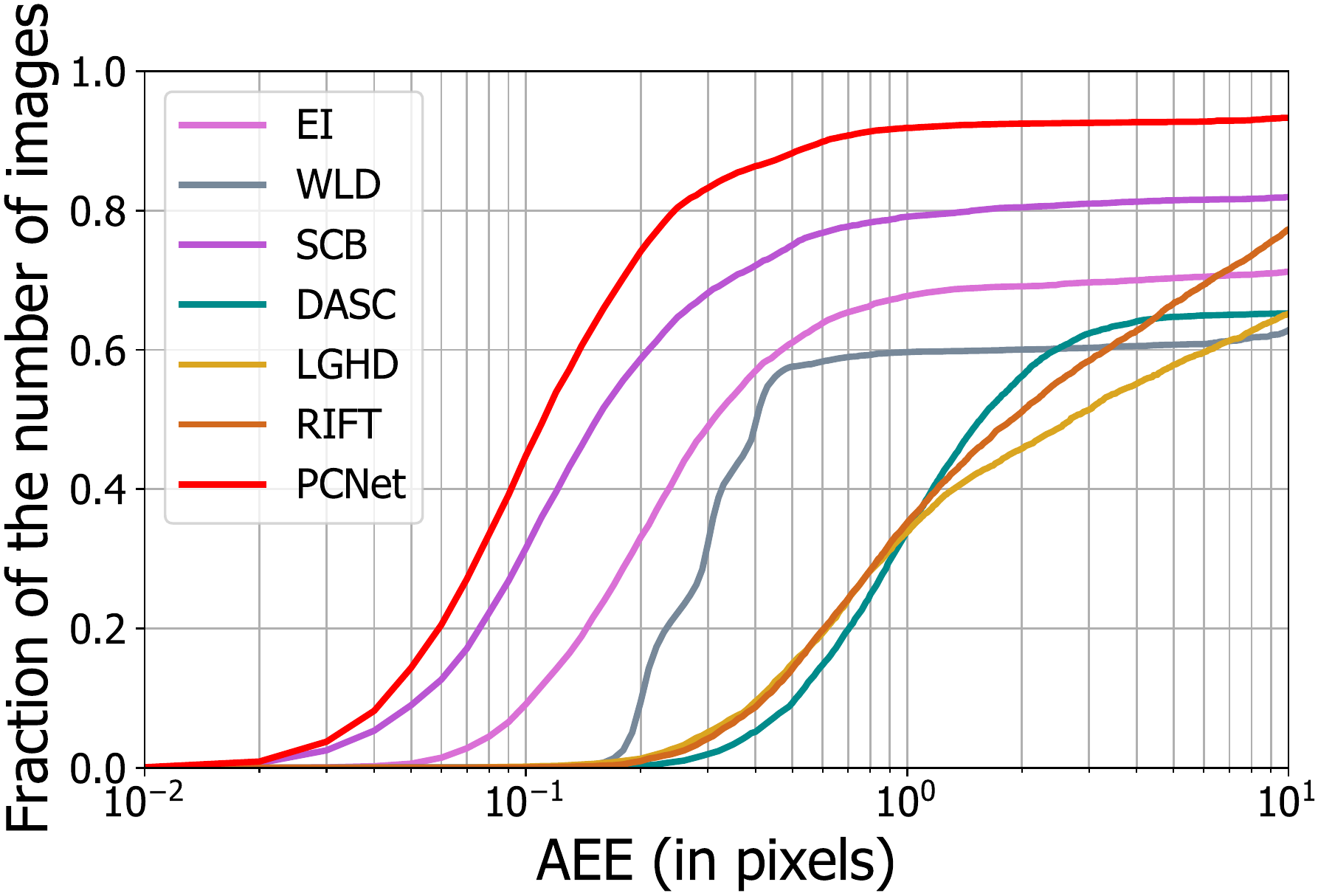}
        \label{fig:harvard_conventional}
    }
    \caption{Registration evaluation on CAVE and Harvard datasets using PCNet and other registration algorithms. The fraction of the number of images within a dataset is plotted with respect to AEE.}
    \label{fig:cave_harvard}
\end{figure*}

Our PCNet is trained once on a subset of the CAVE \cite{yasuma2010generalized} multispectral dataset, and evaluated on various multispectral/multimodal datasets including the CAVE and Harvard \cite{chakrabarti2011statistics} multispectral band image datasets, RGB/NIR \cite{brown2011multi} pairwise multispectral image dataset, and flash/no-flash \cite{he2014saliency} dataset. The training subset for our PCNet is not included in the performance evaluation. We illustrate the above-mentioned datasets in Fig. \ref{fig:cave_harvard} and \ref{fig:rgbnir_flf_exp}. The scenes employed for further illustration are marked with \textit{S1} $\sim$ \textit{S7}. 

The registration performance of our PCNet is compared with other state-of-the-art similarity enhancement algorithms including entropy image (EI) \cite{wachinger2010structural}, WLD \cite{chen2009wld}, and structure consistency boosting (SCB) transform \cite{cao2020boosting}. The same aforementioned hierarchical registration strategy is employed for all the above similarity enhancement algorithms. We also compare our registration framework using PCNet with state-of-the-art feature-based multispectral/multimodal methods including log-Gabor histogram descriptor (LGHD) \cite{aguilera2015lghd}, radiation-variation insensitive feature transform (RIFT) \cite{li2019rift}, and dense adaptive self-correlation (DASC) descriptor \cite{kim2016dasc}. As DASC produces dense 128-channel output similarity enhancement feature maps, we employ the registration strategy provided by its public source code\footnote{https://seungryong.github.io/DASC/}, which registers the feature maps using SIFT flow\cite{liu2008sift}. The output flow is then fed into the estimateGeometricTransform \cite{torr2000mlesac} function in Matlab, producing the affine registration result for a fair comparison. The affine transform of LGHD and RIFT is also obtained by the estimateGeometricTransform function. For a better comparison, we adopt the registration error as the average Euclidean error (AEE) \cite{chen2017normalized, cao2020boosting} between the pixel positions computed by the ground truth transform $\mathbf{a}_{\textrm{gt}}$ and the estimated transform $\mathbf{\hat{a}}$,
\begin{equation}\label{eq:AEE_error}
e_1 = \frac{1}{M}{\sum \limits_{p=1}^M\|\tilde{p}(p,\mathbf{a}_{\textrm{gt}}) - \tilde{p}(p,\mathbf{\hat{a}}) \|_2},
\end{equation}
where $M$ indicates the number of pixels of an image, and $\tilde{p}$ the warped pixel position by applying the transform $\mathbf{a}$ on pixel $p$. 

For an extensive evaluation, we keep the reference image fixed and warp the floating image with the simulated transform of three degrees, including small deformation $\mathbf{a}_{\textrm{gt,s}} = (1.1, 0.1, -10, -0.1, 1.1, 10)^{\textsf{T}}$, middle deformation $\mathbf{a}_{\textrm{gt,m}} = (1.15, 0.15, -15, -0.15, 1.15, 15)^{\textsf{T}}$, and large deformation $\mathbf{a}_{\textrm{gt,l}} = (1.2, 0.2, -20, -0.2, 1.2, 20)^{\textsf{T}}$. In the comparison with the above methods, all degrees of deformations are evaluated on $256 \times 256$ images if not otherwise specified. 

We further conduct a comparison of our registration framework with the deep-learning registration methods including DHN \cite{detone2016deep}, MHN \cite{le2020deep}, and UDHN \cite{zhang2020content}. We also investigate the performance improvement by combining our PCNet with the above deep-learning methods as in \cite{zhao2021deep}. As the deep-learning networks are constructed to produce the displacement of the corner points of an image, the average corner error (ACE) \cite{detone2016deep} is used to conduct the registration performance comparison. ACE is formulated as  
\begin{equation}\label{eq:ACE_error}
e_2 = \frac{1}{4}{\sum \limits_{p=1}^4\|\tilde{p}(p,\mathbf{a}_{\textrm{gt}}) - \tilde{p}(p,\mathbf{\hat{a}}) \|_2},
\end{equation}
where the main difference compared to AEE is the error is only calculated on the 4 corner points of an image instead of all points.
The size of the image to be registered is reduced to $128\times128$ following the design of the deep-learning networks. The translation parameters of the simulated transform are reduced in proportion simultaneously.

We employ two ways to display the registration error. The first one is the table including the mean, median, tri-mean, and the mean of the errors below 25th, 50th, 75th, and 95th percentiles (denoted by best25\%, best50\%, best75\%, and best95\%) within a dataset as in \cite{cao2020boosting}. Another one is the figure that plots the fraction of the number of images with respect to the registration error as in \cite{le2020deep}.

In the following, we first discuss some issues of our PCNet, including the training details and the hyperparameter settings. We then compare our PCNet with the original phase congruency algorithm. Next, we compare our PCNet with the aforementioned registration methods including the conventional ones and deep-learning ones. We also analyze the number of parameters of our PCNet and other deep-learning methods, together with the performance improvement by combining our PCNet with them. Finally, we conduct an ablation study of our PCNet.

\subsection{Training Details and Hyperparameter Setting}\label{subsec:param}

\begin{figure}[!tb]
\centering
\includegraphics[scale=0.7]{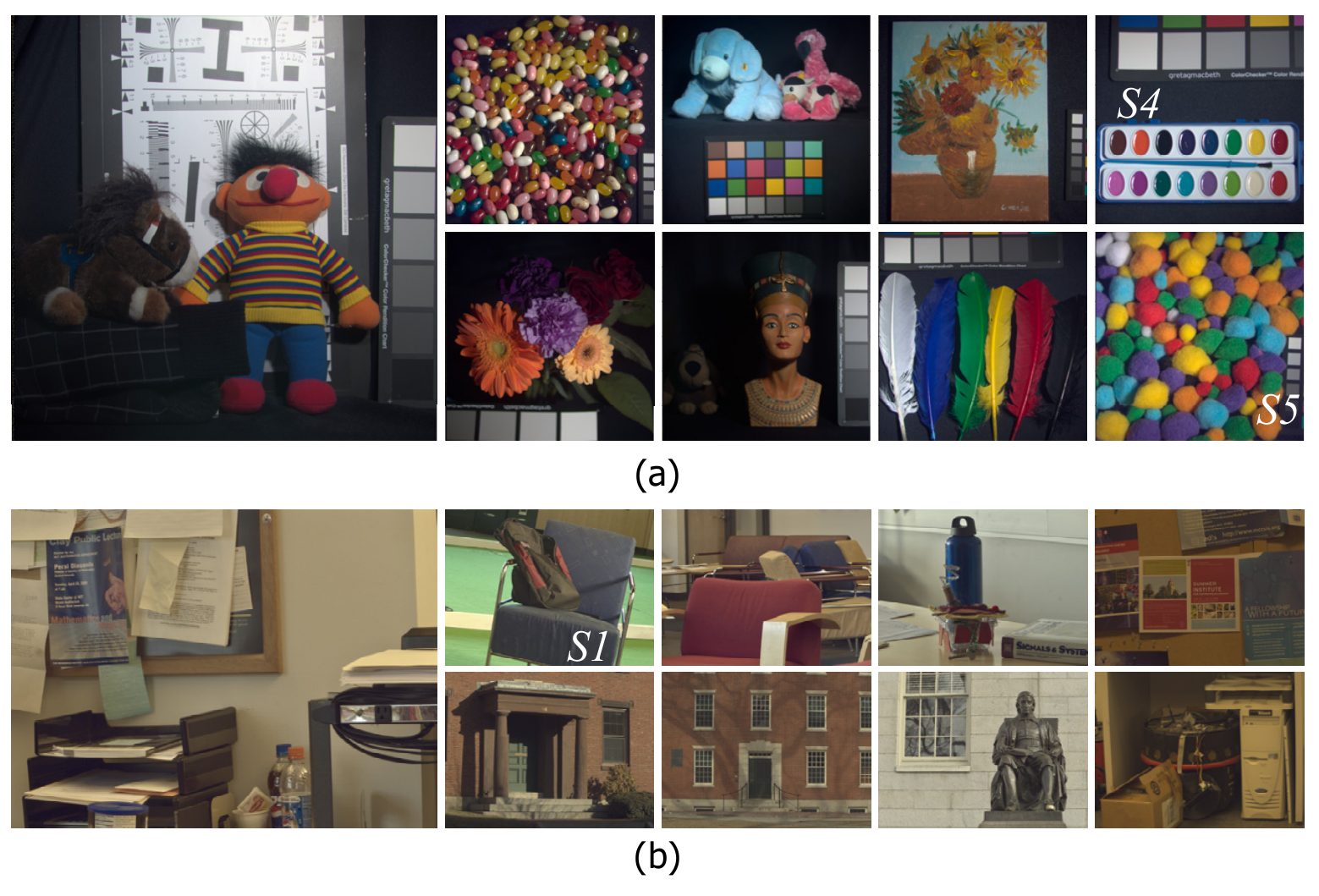}
\caption{Example multispectral images in the visible spectrum (displayed in RGB) from the (a) CAVE and (b) Harvard datasets. The scenes to be used for illustration are marked with \textit{S1}, \textit{S4}, and \textit{S5}.
}
\label{fig:caveharvard_expdata}
\end{figure}

\begin{figure}[!tb]
\centering
\includegraphics[scale=0.45]{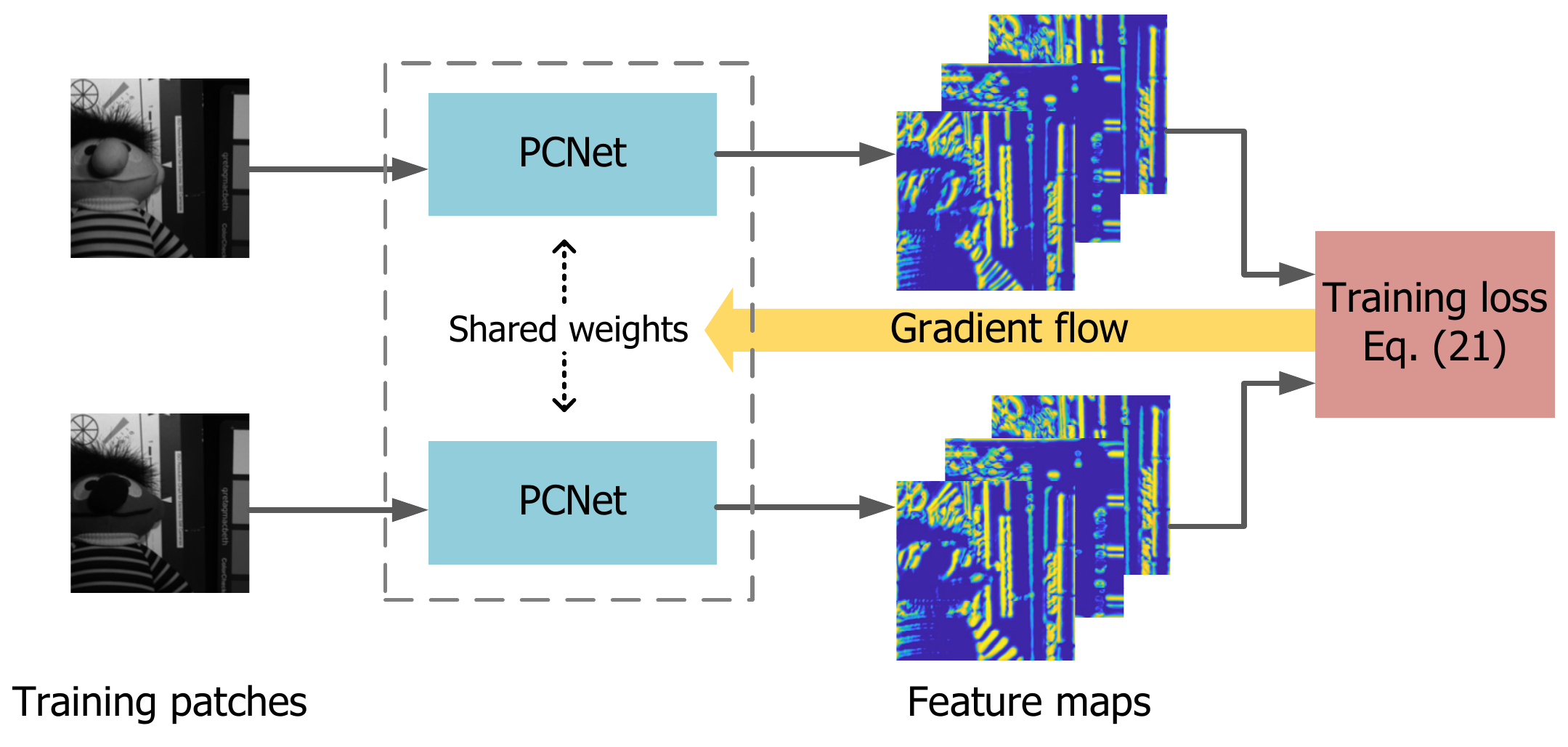}
\caption{The training process of PCNet. The aligned training patches are fed into two identical PCNets of shared weights. The output feature maps of PCNet are evaluated by the training loss, namely equation (\ref{eq:norm_SSIM_loss}). The weights of PCNet are updated by the gradient flow produced by the stochastic gradient descent (SGD) optimizer.}
\label{fig:PCNet_train}
\end{figure}

Our PCNet is trained by a subset of the CAVE \cite{yasuma2010generalized} dataset. The dataset consists of 32 multispectral image scenes, with each including 31 band images ranging from 400 to 700 nm. Example images from the CAVE dataset are displayed in Fig. \ref{fig:caveharvard_expdata}(a). The first 10 scenes in alphabetical order are taken as the training data and the rest as test data. Specifically, we randomly crop several pairs of patches with the size of $200\times200$ for network training. The patches are fed into the network in pair and random order, with data augmentation (brightness, contrast, and saturation adjustment). The training process of PCNet is illustrated in Fig. \ref{fig:PCNet_train}. The input pairwise patches are fed into two identical PCNets of shared weights, and then the normalized structural similarity of output features are evaluated using the training loss Eq. (\ref{eq:norm_SSIM_loss}). The weights of PCNet are updated by the gradient flow produced by the SGD optimizer. Note that our PCNet can be regarded as fully convolutional, which means that the network to work on any size of input image regardless of the patch size for the network training.

\begin{table}[!tb]
\renewcommand\arraystretch{1.2}
\renewcommand\tabcolsep{5pt}
\newcommand{\tabincell}[2]{\begin{tabular}{@{}#1@{}}#2\end{tabular}}
\centering
\caption{Error statistics produced by image registration using PCNet with various degrees of protection parameter for the structural information on the CAVE dataset. For brevity, bestN\% is denoted as B.N\%. The best indicators are in bold.}\label{tab:PCNet_c}
\begin{tabular}{c|c|c|c|c|c|c|c}
  \hline\hline
  \multicolumn{1}{c|}{$c$}&
  \multicolumn{1}{c|}{\textrm{Mean}}&
  \multicolumn{1}{c|}{\textrm{Med.}}&
  \multicolumn{1}{c|}{\textrm{Tri.}}& 
  \multicolumn{1}{c|}{\tabincell{c}{\textrm{B.25\%}}}&
  \multicolumn{1}{c|}{\tabincell{c}{\textrm{B.50\%}}}&
  \multicolumn{1}{c|}{\tabincell{c}{\textrm{B.75\%}}}&
  \multicolumn{1}{c}{\tabincell{c}{\textrm{B.95\%}}}
  \\\hline
  \multicolumn{1}{c|}{0.6} 
  &\multicolumn{1}{c|}{7.89} 
  &\multicolumn{1}{c|}{0.30} 
  &\multicolumn{1}{c|}{0.45}
  &\multicolumn{1}{c|}{0.07} 
  &\multicolumn{1}{c|}{0.14} 
  &\multicolumn{1}{c|}{0.27} 
  &\multicolumn{1}{c}{5.36} 
  \\\hline
  \multicolumn{1}{c|}{0.7} 
  &\multicolumn{1}{c|}{6.55} 
  &\multicolumn{1}{c|}{\textbf{0.23}} 
  &\multicolumn{1}{c|}{\textbf{0.32}}
  &\multicolumn{1}{c|}{\textbf{0.06}} 
  &\multicolumn{1}{c|}{\textbf{0.11}} 
  &\multicolumn{1}{c|}{\textbf{0.20}} 
  &\multicolumn{1}{c}{\textbf{4.13}} 
  \\\hline
  \multicolumn{1}{c|}{0.8} 
  &\multicolumn{1}{c|}{\textbf{6.51}} 
  &\multicolumn{1}{c|}{0.25} 
  &\multicolumn{1}{c|}{0.36} 
  &\multicolumn{1}{c|}{\textbf{0.06}} 
  &\multicolumn{1}{c|}{\textbf{0.11}} 
  &\multicolumn{1}{c|}{0.21} 
  &\multicolumn{1}{c}{4.23}

\\ \hline \hline
\end{tabular}
\end{table}

\begin{table}[!tb]
\renewcommand\arraystretch{1.2}
\renewcommand\tabcolsep{5pt}
\newcommand{\tabincell}[2]{\begin{tabular}{@{}#1@{}}#2\end{tabular}}
\centering
\caption{Error statistics produced by image registration using PCNet with various numbers of orientation of the modified learnable Gabor kernels on the CAVE dataset. For brevity, bestN\% is denoted as B.N\%. The best indicators are in bold.}\label{tab:PCNet_o}
\begin{tabular}{c|c|c|c|c|c|c|c}
  \hline\hline
  \multicolumn{1}{c|}{$N_o$}&
  \multicolumn{1}{c|}{\textrm{Mean}}&
  \multicolumn{1}{c|}{\textrm{Med.}}&
  \multicolumn{1}{c|}{\textrm{Tri.}}& 
  \multicolumn{1}{c|}{\tabincell{c}{\textrm{B.25\%}}}&
  \multicolumn{1}{c|}{\tabincell{c}{\textrm{B.50\%}}}&
  \multicolumn{1}{c|}{\tabincell{c}{\textrm{B.75\%}}}&
  \multicolumn{1}{c}{\tabincell{c}{\textrm{B.95\%}}}
  \\\hline
  \multicolumn{1}{c|}{3} 
  &\multicolumn{1}{c|}{8.34} 
  &\multicolumn{1}{c|}{0.26} 
  &\multicolumn{1}{c|}{0.45}
  &\multicolumn{1}{c|}{0.06} 
  &\multicolumn{1}{c|}{0.12} 
  &\multicolumn{1}{c|}{0.24} 
  &\multicolumn{1}{c}{5.34} 
  \\\hline
  \multicolumn{1}{c|}{6} 
  &\multicolumn{1}{c|}{6.55} 
  &\multicolumn{1}{c|}{\textbf{0.23}} 
  &\multicolumn{1}{c|}{0.32}
  &\multicolumn{1}{c|}{\textbf{0.06}} 
  &\multicolumn{1}{c|}{\textbf{0.11}} 
  &\multicolumn{1}{c|}{\textbf{0.20}} 
  &\multicolumn{1}{c}{4.13} 
  \\\hline
  \multicolumn{1}{c|}{9} 
  &\multicolumn{1}{c|}{\textbf{6.18}} 
  &\multicolumn{1}{c|}{0.24} 
  &\multicolumn{1}{c|}{\textbf{0.31}} 
  &\multicolumn{1}{c|}{\textbf{0.06}} 
  &\multicolumn{1}{c|}{\textbf{0.11}} 
  &\multicolumn{1}{c|}{\textbf{0.20}} 
  &\multicolumn{1}{c}{\textbf{3.81}}

\\ \hline \hline
\end{tabular}
\end{table}

\begin{table*}[!htb]\small
\renewcommand\arraystretch{1.25}
\renewcommand\tabcolsep{1.75pt}
\newcommand{\tabincell}[2]{\begin{tabular}{@{}#1@{}}#2\end{tabular}}
\centering
\caption{Error statistics produced by image registration using PCNet and PC-org with various numbers of orientation of filters on the CAVE dataset. For brevity, bestN\% is denoted as B.N\%. The best indicators are in bold. Note that B.50\% is omitted for a better article layout.}\label{tab:PC_vs_PCNet}
\begin{tabular}{c|c|c|c|c|c|c|c|c|c|c|c|c}
  \hline\hline
  \multicolumn{1}{c|}{}&\multicolumn{6}{c|}{PCNet}&\multicolumn{6}{|c}{PC-org}\\\cline{2-13} \hline
  \multicolumn{1}{c|}{$N_o$}&
  \multicolumn{1}{c|}{\textrm{Mean}}&
  \multicolumn{1}{c|}{\textrm{Med.}}&
  \multicolumn{1}{c|}{\textrm{Tri.}}& 
  \multicolumn{1}{c|}{\tabincell{c}{\textrm{B.25\%}}}&
  \multicolumn{1}{c|}{\tabincell{c}{\textrm{B.75\%}}}&
  \multicolumn{1}{c|}{\tabincell{c}{\textrm{B.95\%}}}&
  \multicolumn{1}{|c|}{\textrm{Mean}}&
  \multicolumn{1}{c|}{\textrm{Med.}}&
  \multicolumn{1}{c|}{\textrm{Tri.}}& 
  \multicolumn{1}{c|}{\tabincell{c}{\textrm{B.25\%}}}&
  \multicolumn{1}{c|}{\tabincell{c}{\textrm{B.75\%}}}&
  \multicolumn{1}{c}{\tabincell{c}{\textrm{B.95\%}}} \\\hline
  
   \multicolumn{1}{c|}{2} 
  &\multicolumn{1}{c|}{\textbf{7.35}} 
  &\multicolumn{1}{c|}{\textbf{0.26}} 
  &\multicolumn{1}{c|}{\textbf{0.42}} 
  &\multicolumn{1}{c|}{0.07} 
  &\multicolumn{1}{c|}{\textbf{0.23}} 
  &\multicolumn{1}{c|}{\textbf{4.90}} 

  &\multicolumn{1}{|c|}{10.49} 
  &\multicolumn{1}{c|}{0.27} 
  &\multicolumn{1}{c|}{4.14} 
  &\multicolumn{1}{c|}{\textbf{0.04}} 
  &\multicolumn{1}{c|}{1.05} 
  &\multicolumn{1}{c}{7.64}
  \\\hline
  \multicolumn{1}{c|}{6} 
  &\multicolumn{1}{c|}{\textbf{6.55}} 
  &\multicolumn{1}{c|}{\textbf{0.23}} 
  &\multicolumn{1}{c|}{\textbf{0.32}} 
  &\multicolumn{1}{c|}{\textbf{0.06}} 
  &\multicolumn{1}{c|}{\textbf{0.20}} 
  &\multicolumn{1}{c|}{\textbf{4.13}}

  &\multicolumn{1}{|c|}{8.73} 
  &\multicolumn{1}{c|}{0.25} 
  &\multicolumn{1}{c|}{0.47} 
  &\multicolumn{1}{c|}{\textbf{0.06}} 
  &\multicolumn{1}{c|}{0.24} 
  &\multicolumn{1}{c}{5.82} 
  \\\hline
  \multicolumn{1}{c|}{9} 
  &\multicolumn{1}{c|}{\textbf{6.18}} 
  &\multicolumn{1}{c|}{\textbf{0.24}} 
  &\multicolumn{1}{c|}{\textbf{0.31}} 
  &\multicolumn{1}{c|}{\textbf{0.06}} 
  &\multicolumn{1}{c|}{\textbf{0.20}} 
  &\multicolumn{1}{c|}{\textbf{3.81}} 

  &\multicolumn{1}{|c|}{8.19} 
  &\multicolumn{1}{c|}{0.27} 
  &\multicolumn{1}{c|}{0.41} 
  &\multicolumn{1}{c|}{0.07} 
  &\multicolumn{1}{c|}{0.24} 
  &\multicolumn{1}{c}{5.33} 
\\ \hline \hline
\end{tabular}
\end{table*}

We set the scales for the phase congruency estimation as $4$, with the scaling factor between successive filters being $2$. For the 4 scales of the modified learnable Gabor kernels, the sizes of the learnable convolutional kernels are set as $7\times7$, $13\times13$, $19\times19$, and $25 \times 25$. It is worth noting that although the above learnable kernels seem to be overly large for common CNNs, they function well under the regularization of Gabor wavelets. The modified learnable Gabor kernels are set to stride 1. For the two trainable layers, we set the initial value of $\alpha = 2$ as it is originally defined as the scaling factor, and we set $\beta = 1$. As for other training details, we adopt the SGD optimizer with an initial learning rate of 0.01 and weight decay of 0.00003. The network training finishes after 25 epochs with a batch size of 100.

We then evaluate two hyperparameters of PCNet, which are the structure protection parameter $c$ in (\ref{eq:norm_SSIM_loss}) and the orientation number of the modified learnable Gabor kernels $N_{o}$ in (\ref{eq:PCNet_multiori}). We evaluate the effect of both parameters in terms of registration accuracy. For the test data of CAVE, we take the 16th band image (i.e. 550 nm) in each scene as the reference image and generate floating images by imposing the aforementioned simulated deformations to all band images. In this manner, we get 2046 image pairs (22 scenes $\times$ 31 bands $\times$ 3 deformations) for registration experiments. 

As elaborated previously, hyperparameter $c$ controls the degree of protection for the structural information, and a larger $c$ means better protection. Nevertheless, if $c$ grows too large, the similarity of the output phase congruency features will be violated. Table \ref{tab:PCNet_c} lists the error statistics of our PCNet under various hyperparameter $c$. We can observe that the table supports the above principle. The best registration performance is obtained at $c=0.7$. 

The hyperparameter $N_{o}$ controls the orientation number of the modified learnable Gabor kernels, and thus determines the channel of the output feature maps. Smaller $N_{o}$ can reduce channel number, which can improve the efficiency of image registration with the side effect of worse accuracy. On the contrary, larger $N_{o}$ is likely to produce better registration performance but the registration efficiency will be sacrificed. Table \ref{tab:PCNet_o} lists the average errors of our PCNet under various hyperparameter $N_{o}$. It is observed that the best registration performance can be achieved at $N_{o} = 9$. However, the improvement of the registration accuracy from $N_{o}=6$ to $N_{o}=9$ is not significant compared to the cost in computation. Therefore, we set the channel number $N_{o} = 6$.

\begin{figure}[!htb]
\centering
\includegraphics[scale=0.35]{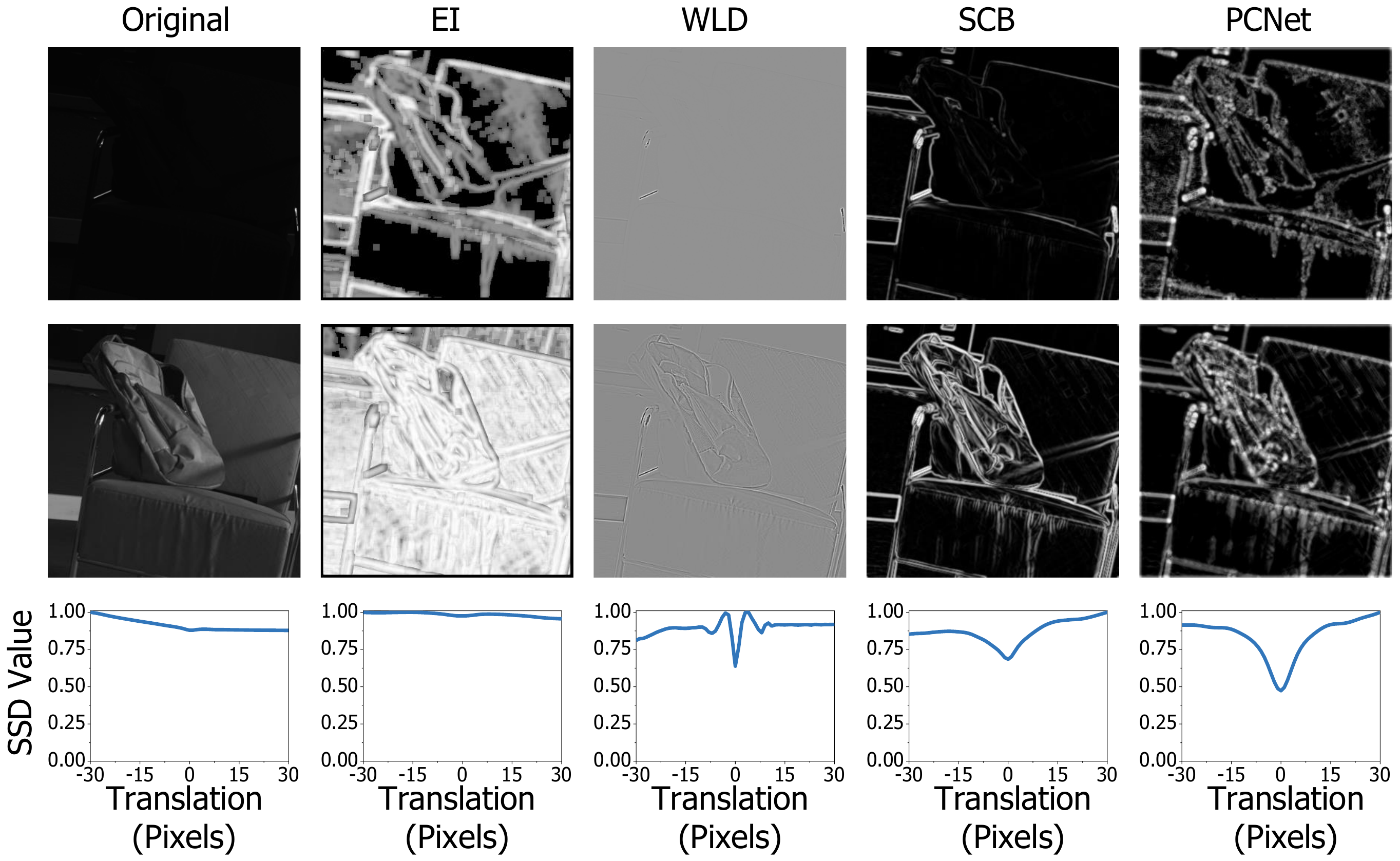}
\caption{Comparison of our PCNet and other similarity enhancement algorithms on scene \textit{S1}. First row: the original and transformed reference images. Second row: the original and transformed floating images. Third row: the SSD distributions with respect to the horizontal translation from -30 to 30 pixels.}
\label{fig:SSD_plot_cmp}
\end{figure}

\subsection{Comparison with Conventional Phase Congruency Algorithm}\label{subsec:cmp_org}

Thanks to the modification strategy, our PCNet can achieve better registration performance than the conventional phase congruency algorithm. We conduct similar registration experiments on the CAVE dataset as in Section \ref{subsec:param}. To conduct an exhaustive comparison for both algorithms, we evaluate their registration performance with various orientation number. We note that, a fewer feature channel means higher registration efficiency in the registration step.

We list the error statistics for both algorithms of orientation number $N_o = 2$, $N_o = 6$ and $N_o = 9$ in Table \ref{tab:PC_vs_PCNet}. The conventional phase congruency algorithm is denoted as PC-org. We can observe that within all feature channels, our PCNet produces considerably better results than PC-org. Furthermore, our PCNet with $2$ channels could achieve higher registration accuracy than the PC-org with $9$ channels, which means $4.5 \times$ improvement of computational efficiency in the registration step. Another interesting phenomenon occurs when we focus on the Best25\% and Best95\% statistics. For our PCNet, the Best25\% and Best95\% decrease as $N_o$ increase, which means the overall accuracy improvement. On the contrary, the Best25\% of PC-org increases as $N_o$ increase, which means the loss of accuracy.

\subsection{Results on Multispectral Band Images}\label{subsec:ms_res}

We evaluate our PCNet with other registration algorithms including EI \cite{zitova2003image}, WLD \cite{chen2009wld}, SCB \cite{cao2020boosting}, DASC \cite{kim2016dasc}, LGHD \cite{aguilera2015lghd}, RIFT \cite{li2019rift} on the CAVE \cite{yasuma2010generalized} and Harvard \cite{chakrabarti2011statistics} datasets. The Harvard dataset contains 77 multispectral images of real-world scenes, each with 31 spectral bands ranging from 420 to 720 nm. The sample images are displayed in Fig. \ref{fig:caveharvard_expdata} (b). Similar to the experiment setting for the CAVE dataset, we again take the 16th band image (i.e., 570 nm) of each Harvard scene as the reference image, and generate the floating images by imposing the simulated transforms $\mathbf{a}_{\textrm{gt,s}}, \mathbf{a}_{\textrm{gt,m}}$, and $\mathbf{a}_{\textrm{gt,l}}$. In this way, we conduct 2046 image pairs (22 scenes $\times$ 31 bands $\times$ 3 deformations) for image registration experiments on the CAVE dataset and 7176 image pairs (77 scenes $\times$ 31 bands $\times$ 3 deformations) on the Harvard dataset. 

We plot the fraction of the number of images with respect to AEE within CAVE and Harvard datasets in Fig. \ref{fig:cave_harvard}. It is observed that our PCNet enjoys the best registration performance on the Harvard dataset, and also keeps almost the best performance on the CAVE dataset except for the slight performance degradation compared to SCB when AEE is lower than 0.07 pixels. As for other competitors, SCB and EI perform relatively better than WLD. As for the feature-based methods, they produce a considerable amount of registration AEE lower than 10 pixels. However, their registration result is not as accurate as PCNet.

\begin{figure*}
    \centering
    \subfigure[Evaluation on flash/no-flash dataset.]
    {
        \centering
        \includegraphics[scale=0.32]{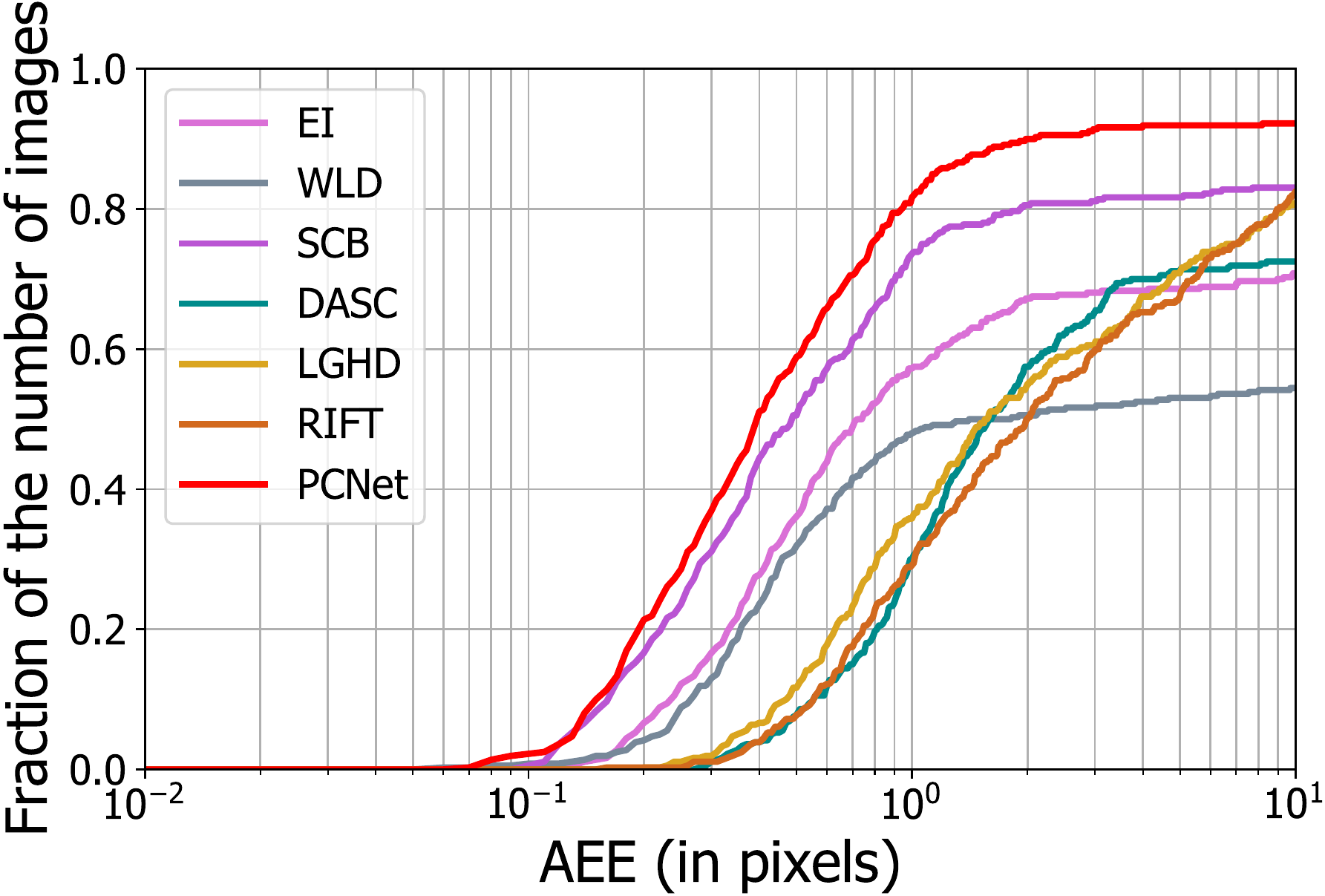}
        \label{fig:flf_conventinonal}
    }
    \subfigure[Evaluation on RGB/NIR dataset.]
    {
        \centering
        \includegraphics[scale=0.32]{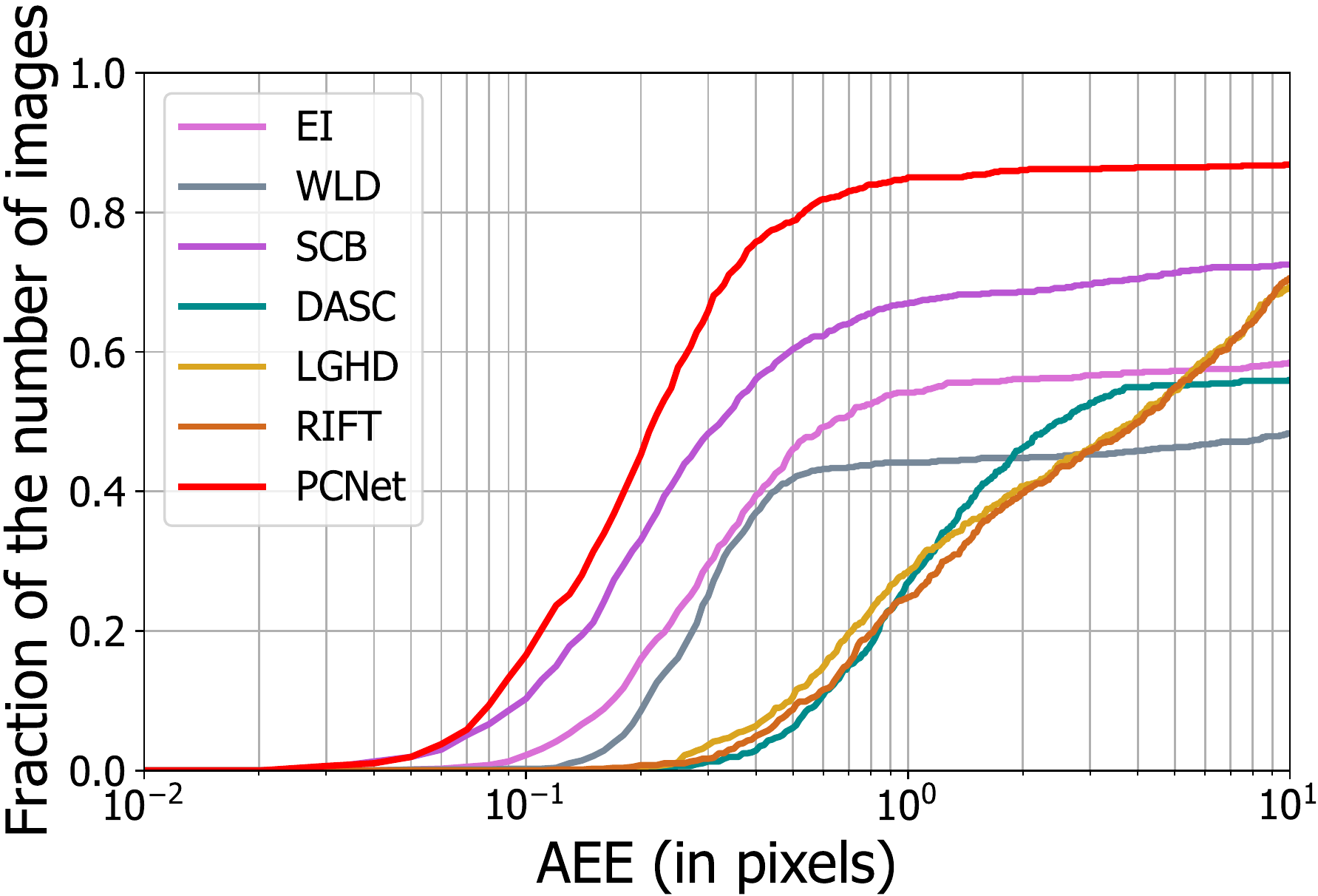}
        \label{fig:rgbnir_conventional}
    }
    \caption{Registration evaluation on flash/no-flash and RGB/NIR datasets using PCNet and other registration algorithms. The fraction of the number of images within a dataset is plotted with respect to AEE.}
    \label{fig:flf_rgbnir}
\end{figure*}

We further demonstrate the similarity enhanced outputs together with the corresponding SSD plots for all the similarity enhancement methods in Fig. \ref{fig:SSD_plot_cmp}. It is observed that the similarity enhanced outputs for EI have significant differences, which results in the SSD plot failing to indicate the best registration position. The similarity enhanced outputs of WLD and SCB are of weak consistency, and hence their SSD plots give weak guidance for the best registration position. In comparison, PCNet produces consistent similarity enhancement outputs, and its SSD plot is of a larger capture range and stronger minimum peak than any other algorithms.

\subsection{Results on Flash/no-flash and RGB/NIR Image Pairs}

\begin{figure*}[!tb]
\centering
\includegraphics[scale=0.7]{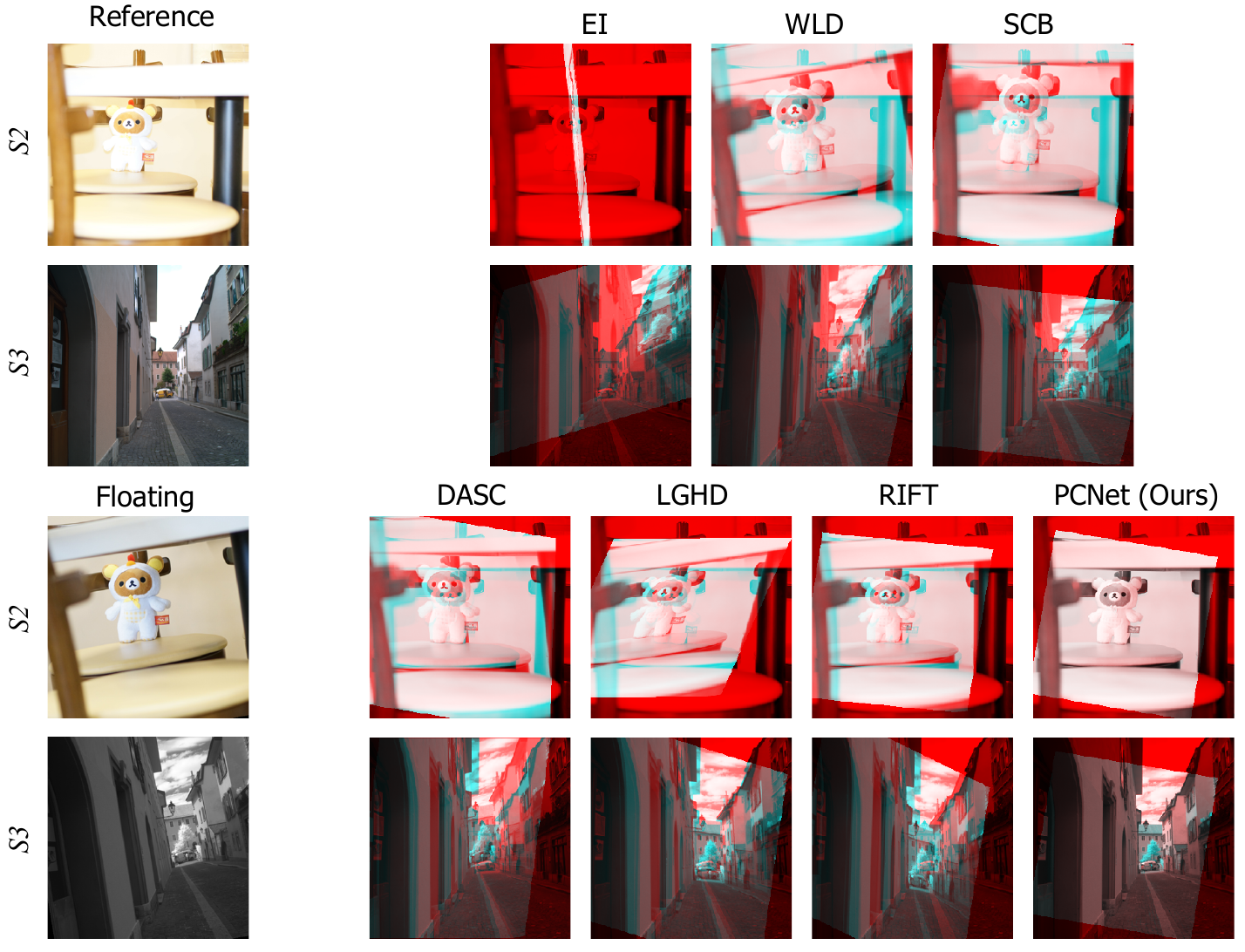}
\caption{Registration results produced by PCNet and other registration algorithms, under the large degree of transform. First row: registration results of RGB/NIR image pair, scene \textit{S2}. Second row: registration results of flash/no-flash image pairs, scene \textit{S3}.}
\label{fig:rgbnir_flf_reg_res}
\end{figure*}

We further evaluate our PCNet together with other registration methods on flash/no-flash and RGB/NIR datasets. The flash/no-flash dataset \cite{he2014saliency} includes 120 indoor and outdoor image pairs. The RGB/NIR dataset \cite{brown2011multi} contains 256 RGB/NIR image pairs of various categories of scenes. The example images from the above datasets are displayed in Fig. \ref{fig:rgbnir_flf_exp}. For the flash/no-flash dataset, we set the flash image as the reference image and no-flash as the floating one. For the RGB/NIR dataset, we set the RGB image as the reference image and NIR as the floating one. The deformations imposed on the floating images are the same as in previous experiments. Totally we conduct 360 experiments (120 scenes $\times$ 3 deformations) on the flash/no-flash dataset and 768 experiments (256 scenes $\times$ 3 deformations) on the RGB/NIR dataset. It is worth noting that though our PCNet is trained on a subset of CAVE multispectral band data, experimental results show that it performs well on flash/no-flash and RGB/NIR datasets without any retraining. 

We draw the fraction of the number of images with respect to AEE within flash/no-flash and RGB/NIR datasets in Fig. \ref{fig:flf_rgbnir}. It is observed that our PCNet still outperforms other competitors without any retraining, which means a superior generalization ability for different image data. As for other competitors, they produce similar performance as in the previous subsection \ref{subsec:ms_res}.

Fig. \ref{fig:rgbnir_flf_reg_res} displays the registration results of scenes \textit{S2} and \textit{S3} produced by PCNet and other registration algorithms. The registration results are illustrated by overlapping the reference image and the registered floating image and then displaying them in false color. It is observed that for both scenes, only PCNet produces registration results stably and accurately. It is worth noting that LGHD and RIFT generally produce roughly right registration results yet lack precision. 

\subsection{Comparison with Deep-Learning Methods}

In this subsection, we compare our PCNet with 3 state-of-the-art registration algorithms using deep CNNs, including DHN \cite{detone2016deep}, MHN \cite{le2020deep}, and UDHN \cite{zhang2020content}. The above networks are retrained on the first 10 scenes in alphabetic order of the CAVE dataset and evaluated on the rest of the dataset, which is denoted as CAVE-CAVE. We then evaluate all the above methods on RGB/NIR dataset without retraining, which is denoted as CAVE-RGB/NIR, to compare their generalization ability. We note that our PCNet is only trained once on the first 10 scenes in alphabetic order of the CAVE dataset for a fair comparison. 

\begin{figure*}
    \centering
    \subfigure[CAVE-CAVE evaluation with deep-learning methods.]
    {
        \centering
        \includegraphics[scale=0.32]{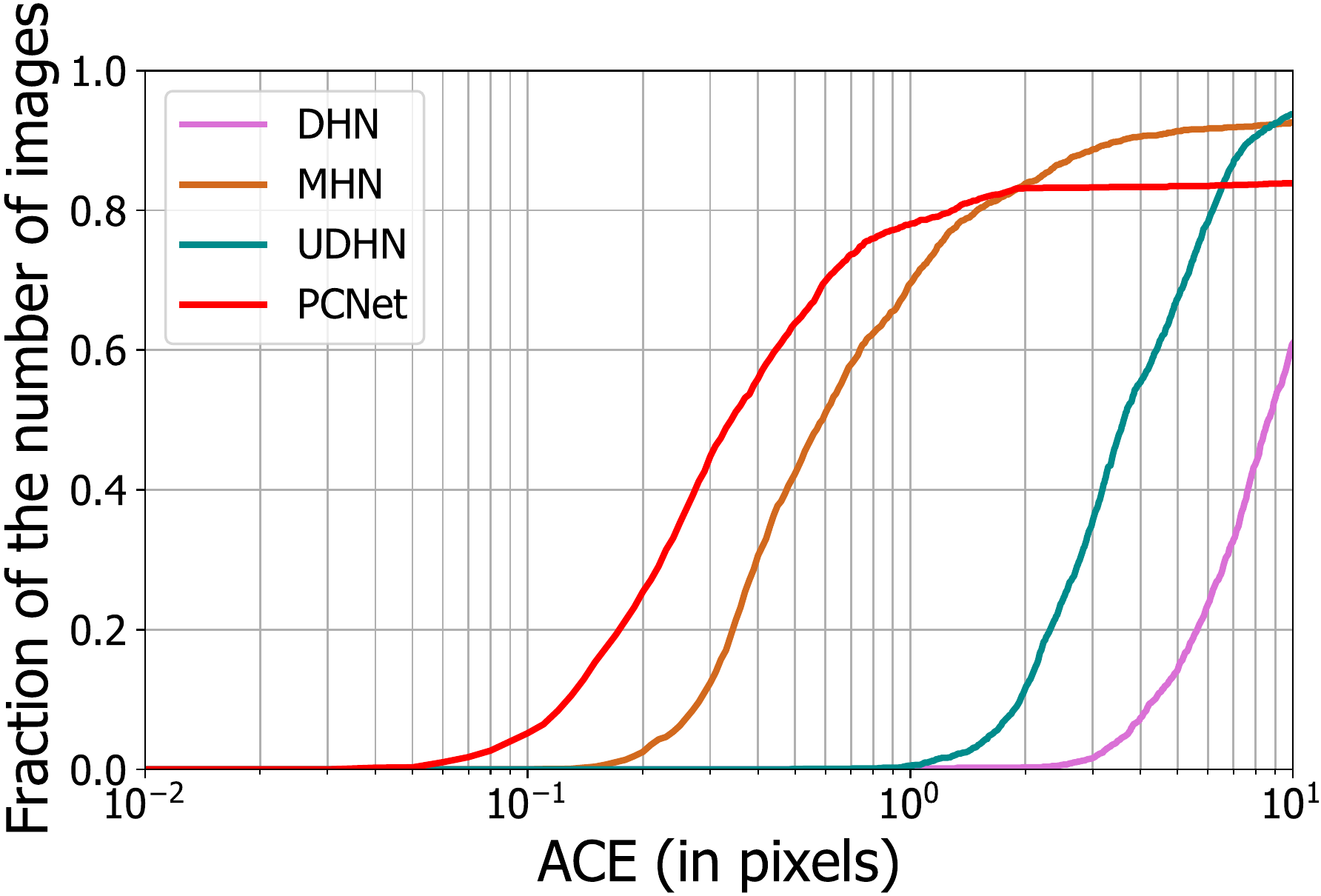}
        \label{fig:cave_deep}
    }
    \subfigure[CAVE-RGB/NIR evaluation with deep-learning methods.]
    {
        \centering
        \includegraphics[scale=0.32]{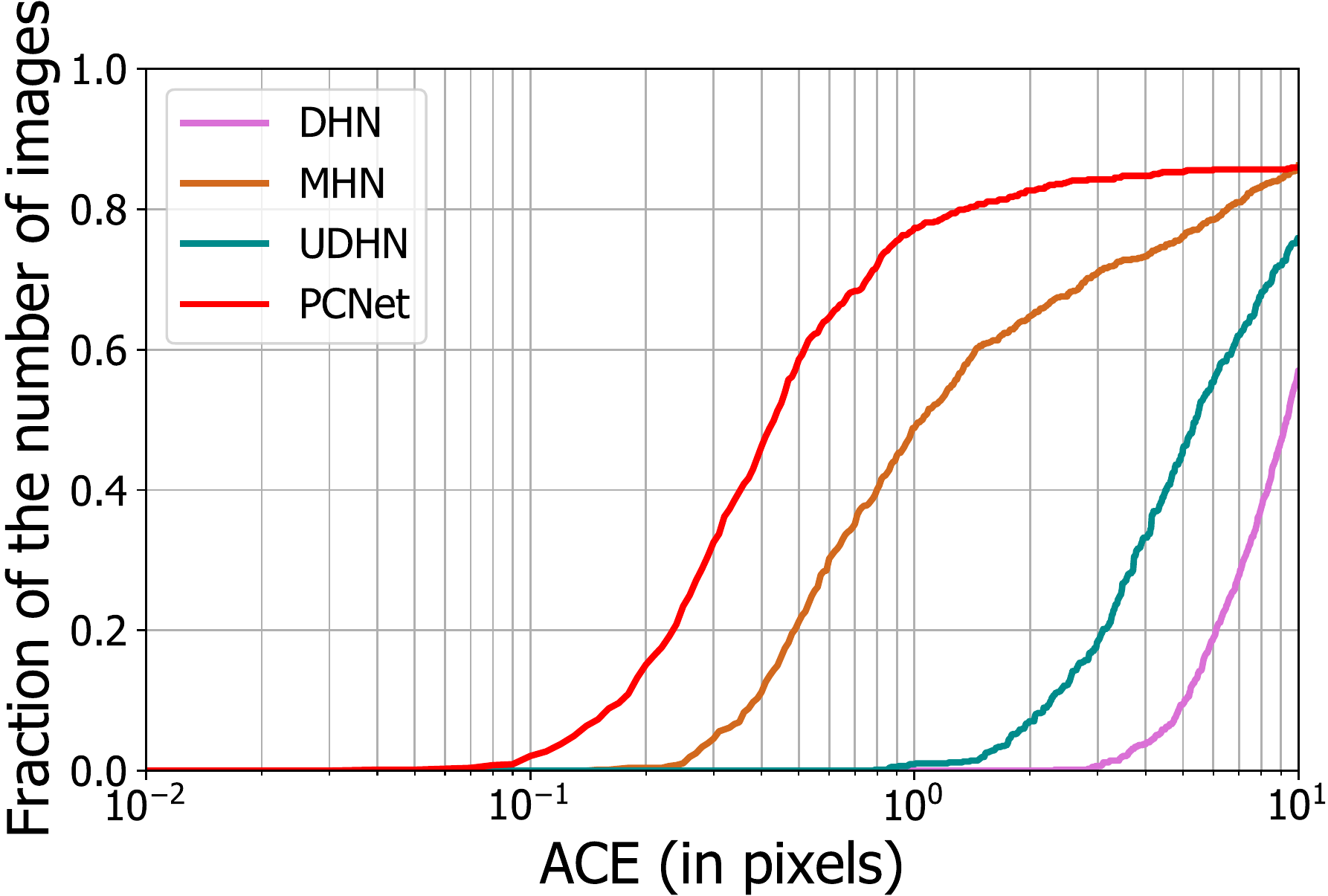}
        \label{fig:rgbnir_deep}
    }
    \caption{Registration evaluation on CAVE and RGB/NIR datasets using PCNet and other deep-learning registration algorithms. The fraction of the number of images within a dataset is plotted with respect to ACE.}
    \label{fig:cave_rgbnir_deep}
\end{figure*}

\begin{figure*}
    \centering
    \subfigure[CAVE-CAVE evaluation of PCNet boosted deep-learning methods.]
    {
        \centering
        \includegraphics[scale=0.32]{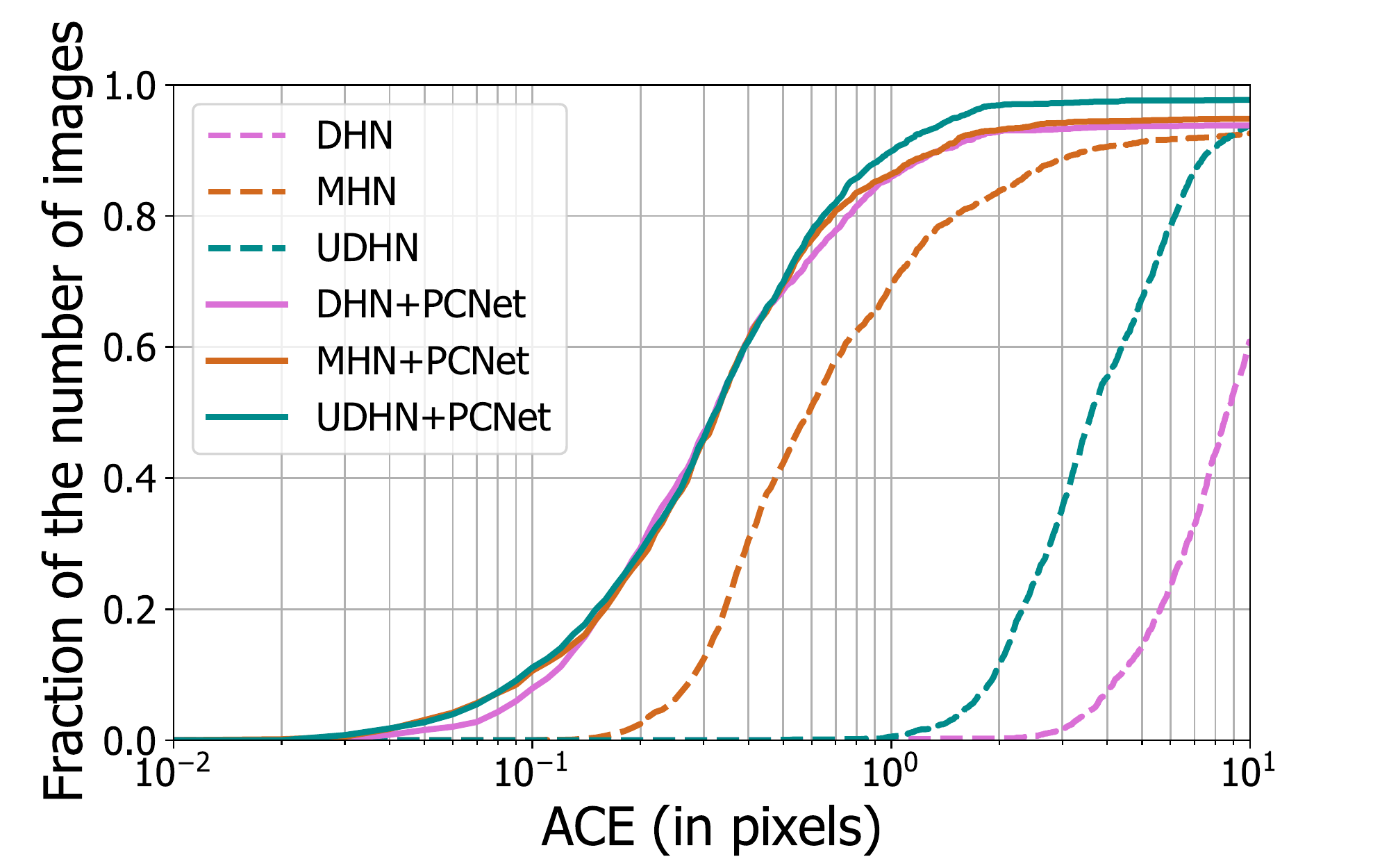}
        \label{fig:cave_deeppcnet}
    }
    \subfigure[CAVE-RGB/NIR evaluation of PCNet boosted deep-learning methods.]
    {
        \centering
        \includegraphics[scale=0.32]{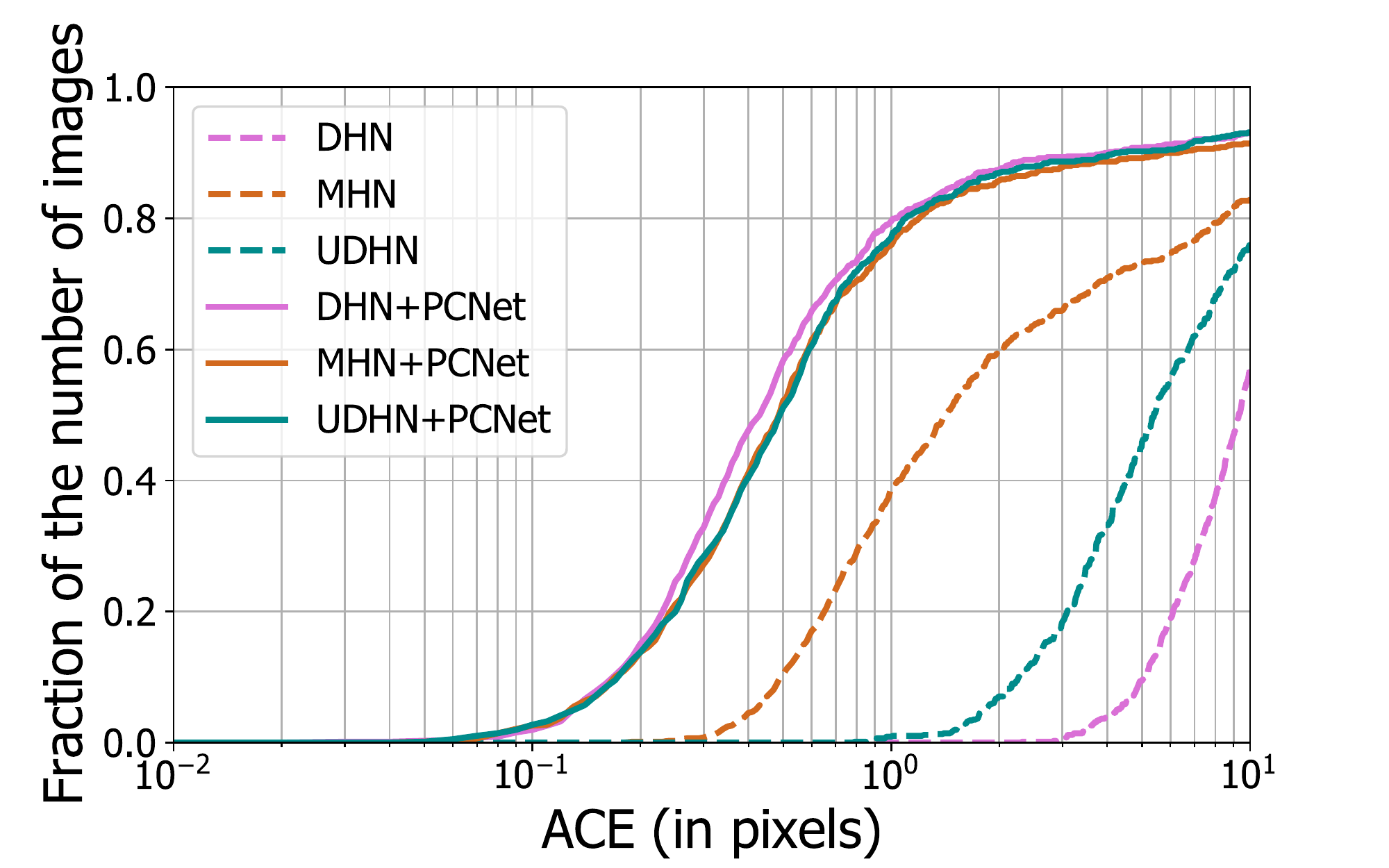}
        \label{fig:rgbnir_deeppcnet}
    }
    \caption{Registration evaluation on CAVE and RGB/NIR datasets using deep-learning registration algorithms and their PCNet boosted versions. The fraction of the number of images within a dataset is plotted with respect to ACE.}
    \label{fig:deeppcnet}
\end{figure*}

\begin{figure*}[!tb]
\centering
\includegraphics[scale=0.3]{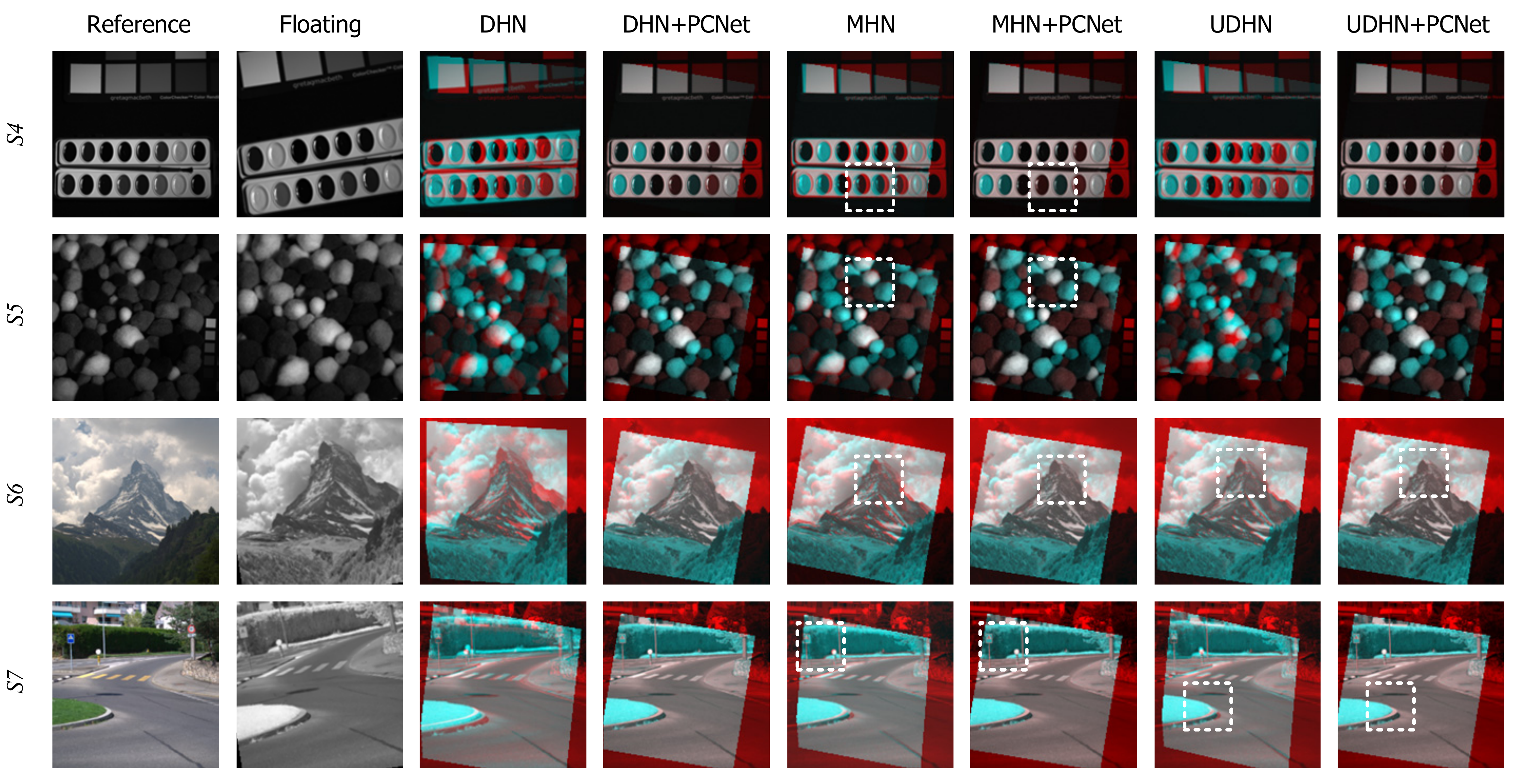}
\caption{Registration results produced by deep-learning registration algorithms together with their PCNet boosted version. First and second row: registration results of CAVE image pair, scene \textit{S4} and \textit{S5}. Third and fourth row: registration results of RGB/NIR image pairs, scene \textit{S6} and \textit{S7}. The white boxes highlight the details for comparison.}
\label{fig:deep_reg_res}
\end{figure*}

We draw the fraction of the number of images with respect to ACE within CAVE and RGB/NIR datasets in Fig. \ref{fig:cave_rgbnir_deep}. It is observed that in the CAVE-CAVE evaluation in Fig. \ref{fig:cave_deep}, while our PCNet achieves the largest proportion when ACE is less than around 2 pixels, it is surpassed by MHN and UDHN otherwise. The strong learning ability of deep neural networks contributes to the outstanding performance of MHN and UDHN. We then focus on the cross dataset evaluation, namely the CAVE-RGB/NIR evaluation in Fig. \ref{fig:rgbnir_deep}. It is observed that PCNet outperforms all competitors, which indicates a relatively better generalization ability of the registration framework using PCNet. 

An interesting phenomenon is observed in Fig. \ref{fig:cave_deep} that our proposed PCNet registration framework can achieve relatively higher accuracy than the deep-learning methods as it keeps the largest proportion when ACE is less than around 2 pixels. On the contrary, in the case that ACE ranges from 2 pixels to 10 pixels, the deep-learning methods such as MHN and UDHN perform better. The advantages of both methods can be combined by using our PCNet to boost the registration produced by the deep-learning methods as in \cite{zhao2021deep}. 

We plot the boosted and the original results on CAVE and RGB/NIR datasets of deep-learning methods in Fig. \ref{fig:deeppcnet}. It is observed that our PCNet can significantly boost the registration accuracy of all the deep-learning methods by a large gap. The obvious registration performance gap among the deep-learning methods (e.g. DHN and MHN) is removed after the boosting of our PCNet, which reveals the capacity of our PCNet in another aspect. It is worth noting that as is explored in \cite{le2020deep}, a more cascading of deep neural networks is unable to boost the registration accuracy. 

We further list the the percentage of number of images under ACEs of $1$, $5$, and $10$ pixels for each deep-learning method in Table \ref{tab:deep_percent_cpr}. It is observed that our PCNet boosts the registration accuracy of the deep-learning methods via a considerably large gap. On the CAVE dataset, our PCNet raises the percentage under $1$ pixel of UDHN from 0.2\% to 89.9\%. On the RGB/NIR dataset, PCNet boosts the percentage under $1$ pixel of DHN from 0.0\% to 79.6\%.

\begin{table}[!htb]
\renewcommand\arraystretch{1.2}
\renewcommand\tabcolsep{3pt}
\centering
\caption{The percentage of number of images under different degrees of ACEs using deep-learning methods and their PCNet boosted versions.}\label{tab:deep_percent_cpr}
\begin{tabular}{c|ccc|ccc}
  \hline\hline
  \multicolumn{1}{c|}{}&\multicolumn{3}{c|}{CAVE}&\multicolumn{3}{c}{RGB/NIR}\\\cline{2-7} \hline
   ACE (pixels) &$<1$ & $<5$ & $<10$&$<1$ & $<5$ & $<10$ 
   \\\hline
  DHN &0.0\% &7.8\% &47.6\%&0.0\% &4.2\% &39.6\%
  \\
  DHN+PCNet &\textbf{86.00\%} &\textbf{93.7\%} &\textbf{93.8\%}&\textbf{79.6\%} &\textbf{90.8\%} &\textbf{93.1\%}
  \\\hline
  MHN &59.8\% &90.0\% &91.9\%&38.9\% &73.2\% &82.8\%
  \\
  MHN+PCNet &\textbf{86.4\%} &\textbf{94.5\%} &\textbf{94.8\%}&\textbf{75.9\%} &\textbf{89.2\%} &\textbf{91.4\%}
  \\\hline
  UDHN &0.2\% &51.0\% &89.8\%&0.1\% &35.3\% &71.0\%
  \\
  UDHN+PCNet &\textbf{89.9\%} &\textbf{97.6\%} &\textbf{97.7\%}&\textbf{77.1\%} &\textbf{90.2\%} &\textbf{93.1\%}
   \\\hline
  \hline
\end{tabular}
\end{table}

Fig. \ref{fig:deep_reg_res} illustrates the registration results of scenes \textit{S4}, \textit{S5}, \textit{S6}, and \textit{S7} produced by deep-learning registration algorithms and their PCNet boosted versions. The white boxes highlight the details for comparison. It is observed that our PCNet can boost the varying degrees of registration results of deep-learning methods significantly. The obvious registration performance gap among the deep-learning methods (e.g. DHN and MHN) is removed after the boosting of our PCNet.

It is worth taking a comparison of the number of parameters for the deep-learning methods and PCNet. It is observed from Table \ref{tab:num_para} that the number of parameters of our PCNet is significantly less than deep-learning algorithms. The number of parameters of PCNet is about $2400\times$ less compared to DHN and $180\times$ to MHN. This indicates that our PCNet can achieve accurate image registration with an extremely lower computation power requirement.

\begin{table}[!htb]
\renewcommand\arraystretch{1.2}
\renewcommand\tabcolsep{3pt}
\centering
\caption{The number of parameters of PCNet and other deep-learning registration algorithms.}\label{tab:num_para}
\begin{tabular}{c|c|c|c|c}
  \hline\hline
   &DHN \cite{detone2016deep}  & MHN \cite{le2020deep} & UDHN \cite{zhang2020content} & PCNet
   \\\hline
  \# Parameters &$ 3.4\times10^7$  & $ 2.6\times10^6$ & $ 2.1\times10^7$ & $ 1.4\times10^4$
   \\\hline
  \hline
\end{tabular}
\end{table}

\subsection{Ablation Study}
We conduct the ablation study of our PCNet for the noise estimation layer (layer1), modified phase deviation estimation layer (layer2), learnable convolutional kernels (LCK), modification of Gabor filter, and Gabor filter. Table \ref{tab:ablation} lists the error statistics of our full PCNet, as well as the networks whose network parts are individually frozen or removed. It is observed that our full PCNet has the best registration performance. Specifically, the networks without the learnable convolutional kernel (LCK), the Gabor filter, or the modification of the Gabor filter are of evident performance degradation, which means the added network parts contribute a lot to the registration accuracy. The freezing of the modified phase deviation estimation layer (layer2) also causes a relatively large drawback in registration accuracy as its tri-mean becomes 2 times the full PCNet.

\begin{figure}[!htb]
\centering
\includegraphics[scale=0.4]{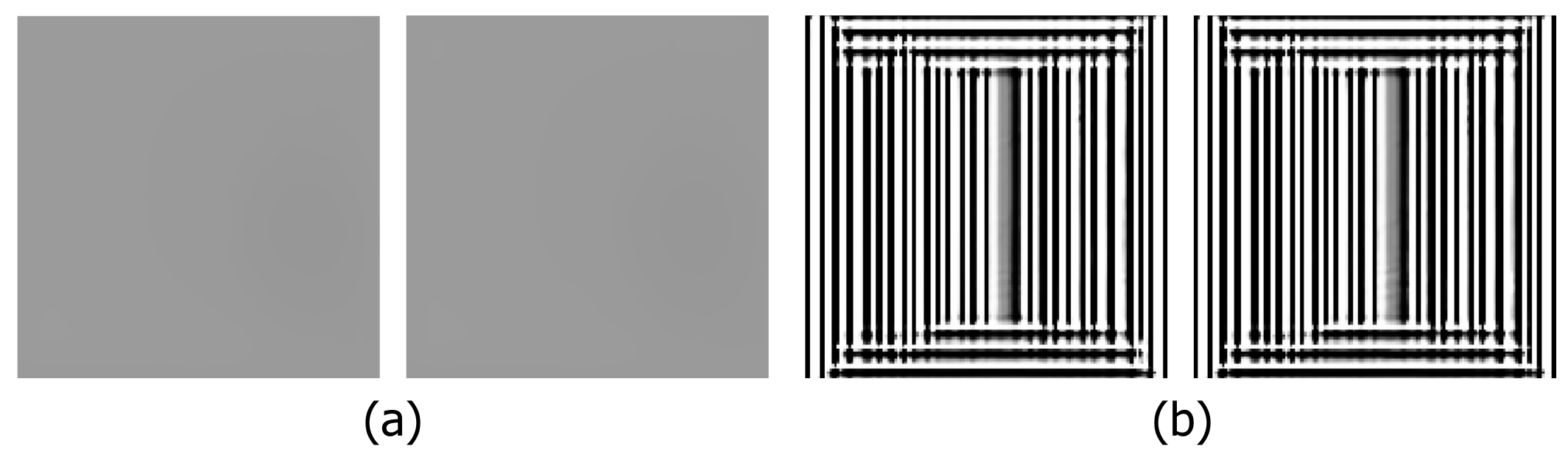}
\caption{Similarity enhanced feature maps generated by U-net. (a) structure protection parameter $c=0.7$. (b) structure protection parameter $c=2$.}
\label{fig:ablation}
\end{figure}

We further validate the necessity of phase congruency in the similarity enhancement task. We replace our PCNet with a U-net \cite{ronneberger2015u}, and then train the network on the same loss Eq. (\ref{eq:norm_SSIM_loss}) as PCNet. However, during training, the output feature maps of U-net using each of the above losses become meaningless patterns that lack information relating to the input images and cannot be used for image registration as illustrated in Fig. \ref{fig:ablation}. It seems that without prior knowledge, it is unlikely to train the similarity enhancement network with the only similarity loss. The same conclusion has also been verified recently. In \cite{zhao2021deep}, the convolutional networks for similarity enhancement are trained by not only constraining the output feature maps to be similar using MSE loss, but also adding a convergence loss to guarantee the output feature maps have a smooth surface around the ground truth transform.

\begin{table}[!tb]\label{tab:ablation}
\renewcommand\arraystretch{1.2}
\renewcommand\tabcolsep{3.75pt}
\newcommand{\tabincell}[2]{\begin{tabular}{@{}#1@{}}#2\end{tabular}}
\centering
\caption{Error statistics produced by image registration using different settings of PCNet on the CAVE dataset. For brevity bestN\% is denoted as B.N\%. The best indicators are in bold.}\label{tab:ablation}
\begin{tabular}{c|c|c|c|c|c|c|c}
  \hline\hline
  \multicolumn{1}{c|}{Ablation part}&
  \multicolumn{1}{c|}{\textrm{Mean}}&
  \multicolumn{1}{c|}{\textrm{Med.}}&
  \multicolumn{1}{c|}{\textrm{Tri.}}& 
  \multicolumn{1}{c|}{\tabincell{c}{\textrm{B.25\%}}}&
  \multicolumn{1}{c|}{\tabincell{c}{\textrm{B.50\%}}}&
  \multicolumn{1}{c|}{\tabincell{c}{\textrm{B.75\%}}}&
  \multicolumn{1}{c}{\tabincell{c}{\textrm{B.95\%}}}
  \\\hline
  \multicolumn{1}{c|}{Freezing layer1} 
  &\multicolumn{1}{c|}{6.86} 
  &\multicolumn{1}{c|}{0.23} 
  &\multicolumn{1}{c|}{0.34}
  &\multicolumn{1}{c|}{\textbf{0.06}} 
  &\multicolumn{1}{c|}{0.11} 
  &\multicolumn{1}{c|}{\textbf{0.20}} 
  &\multicolumn{1}{c}{4.36} 
  \\\hline
  \multicolumn{1}{c|}{Freezing layer2} 
  &\multicolumn{1}{c|}{6.71} 
  &\multicolumn{1}{c|}{0.28} 
  &\multicolumn{1}{c|}{0.62}
  &\multicolumn{1}{c|}{\textbf{0.06}} 
  &\multicolumn{1}{c|}{0.12} 
  &\multicolumn{1}{c|}{0.26} 
  &\multicolumn{1}{c}{4.69} 
  \\\hline
  \multicolumn{1}{c|}{w/o LCK} 
  &\multicolumn{1}{c|}{8.12} 
  &\multicolumn{1}{c|}{0.25} 
  &\multicolumn{1}{c|}{3.30} 
  &\multicolumn{1}{c|}{\textbf{0.06}} 
  &\multicolumn{1}{c|}{0.11} 
  &\multicolumn{1}{c|}{0.28} 
  &\multicolumn{1}{c}{5.84}
    \\\hline
  \multicolumn{1}{c|}{w/o modification} 
  &\multicolumn{1}{c|}{8.51} 
  &\multicolumn{1}{c|}{0.23} 
  &\multicolumn{1}{c|}{3.77} 
  &\multicolumn{1}{c|}{\textbf{0.06}} 
  &\multicolumn{1}{c|}{\textbf{0.10}} 
  &\multicolumn{1}{c|}{0.45} 
  &\multicolumn{1}{c}{6.03}
    \\\hline
  \multicolumn{1}{c|}{w/o Gabor filter} 
  &\multicolumn{1}{c|}{16.06} 
  &\multicolumn{1}{c|}{2.42} 
  &\multicolumn{1}{c|}{8.89} 
  &\multicolumn{1}{c|}{0.39} 
  &\multicolumn{1}{c|}{0.76} 
  &\multicolumn{1}{c|}{6.06} 
  &\multicolumn{1}{c}{13.24}
    \\\hline
  \multicolumn{1}{c|}{Full} 
  &\multicolumn{1}{c|}{\textbf{6.55}} 
  &\multicolumn{1}{c|}{\textbf{0.23}} 
  &\multicolumn{1}{c|}{\textbf{0.32}}
  &\multicolumn{1}{c|}{\textbf{0.06}} 
  &\multicolumn{1}{c|}{0.11} 
  &\multicolumn{1}{c|}{\textbf{0.20}} 
  &\multicolumn{1}{c}{\textbf{4.13}} 

\\ \hline \hline
\end{tabular}
\end{table}

\section{Conclusions}\label{sec:conclusion}

In this paper, we have proposed a network called PCNet to enhance the structure similarity for the purpose of image registration. The prior knowledge of our PCNet is based on phase congruency. PCNet is concise and easy to train thanks to the prior information. It produces satisfactory registration results on a variety of multispectral and multimodal datasets though only trained on a subset of the CAVE multispectral dataset. The PCNet performs better than other state-of-the-art similarity enhancement algorithms and feature-based registration algorithms.

We note that currently our PCNet is directly combined with a traditional registration framework. This direct combination makes the trainable part of the network lightweight but may restrain the registration performance. It would be our future work to transform the current PCNet into a flexible structure that can be easily incorporated into trainable deep-learning registration networks. Considering that in the real world multispectral and multimodal images are usually unregistered, it is also worthwhile to explore the possibility of training PCNet on unregistered images.



 \bibliographystyle{elsarticle-num} 
 \bibliography{REF.bib}






\end{document}